\documentclass[10.5pt,onecolumn]{article}

\usepackage[a4paper,total={6.in, 8in}]{geometry}

\usepackage{hyperref}       
\usepackage{url}            
\usepackage{booktabs}       
\usepackage{amsfonts}       
\usepackage{microtype}      
\usepackage{bm,amsmath,amsthm}

\newtheorem{proposition}{Proposition}
\newtheorem{lemma}{Lemma}

\newtheorem{theorem}{Theorem}

\newtheorem{assumption}{Assumption}
\newtheorem{remark}{Remark}

\allowdisplaybreaks

\usepackage{amstext,amsxtra,amssymb}
\usepackage{accents,color}

\usepackage{setspace}

\makeatletter
\def\wbar{\accentset{{\cc@style\underline{\mskip8mu}}}}
\makeatother

\begin{document}
	
	\title{\textbf{A Multistep Lyapunov Approach for Finite-Time Analysis of Biased Stochastic Approximation}
	}

	\author{
		Gang Wang, Bingcong Li, and Georgios B. Giannakis \thanks{The work was supported partially by NSF grants  1505970, 1711471, and 1901134. The authors are with the Digital Technology Center and the Department of Electrical and Computer Engineering, University of Minnesota, Minneapolis, MN 55455, USA. (e-mail: gangwang@umn.edu; lixx5599@umn.edu; georgios@umn.edu).}
	}

\maketitle

\begin{abstract}
Motivated by the widespread use of temporal-difference (TD-) and Q-learning algorithms in reinforcement learning, this paper studies a class of biased stochastic approximation (SA) procedures under a mild ``ergodic-like'' assumption on the underlying stochastic noise sequence. Building upon a carefully designed \emph{multistep Lyapunov function} that looks ahead to several future updates to accommodate the stochastic perturbations (for control of the gradient bias), we prove a general result on the convergence of the iterates, and use it to derive non-asymptotic bounds on the mean-square error in the case of constant stepsizes. This novel looking-ahead viewpoint renders finite-time analysis of biased SA algorithms under a large family of stochastic perturbations possible. For direct comparison with existing contributions, we also demonstrate these bounds by applying them to TD- and Q-learning with linear function approximation, under the practical Markov chain observation model. The resultant finite-time error bound for both the TD- as well as the Q-learning algorithms is the first of its kind, in the sense that it holds i) for the unmodified versions (i.e., without making any modifications to the parameter updates) using even nonlinear function approximators; as well as for Markov chains ii) under general mixing conditions and iii) starting from any initial distribution, at least one of which has to be violated for existing results to be applicable.  
\end{abstract}

%
%
%
	
	\section{Introduction}
	\label{sec:intr}
	
	Stochastic approximation (SA) algorithms nowadays are widely used in numerous areas, including statistical signal processing, communications, control, optimization, data science, machine learning, and (deep) reinforcement learning (RL). Ever since the seminal contribution \cite{1951sa}, there have been a multitude of efforts on SA schemes, their applications, and theoretical developments; see, for instance, \cite{book2003sa}, \cite{book2009sa}, \cite{siam2009sa}, \cite{nips2011sa}. On the theory side, conventional SA convergence analysis and error bounds are mostly asymptotic---that hold only in the limit as the number of iterations increases to infinity. Nevertheless, recent research efforts have gradually shifted toward developing non-asymptotic performance guarantees---that hold even for finite iterations---for SA algorithms in different settings \cite{siam2009sa}, \cite{nips2011sa}, \cite{colt2019biased}, mainly motivated by the emerging need for dealing with massive data examples in modern large-scale optimization and statistical learning tasks.
	
	Many stochastic control tasks
	 can be naturally formulated as Markov decision processes (MDPs), which provide a flexible framework for modeling decision making in scenarios where outcomes are partly random and partly under the control of a decision maker. 
	 RL is a collection of techniques for solving MDPs, especially when the underlying transition mechanism is unknown  \cite{1988td},
	\cite{thesis1989}, 
	\cite{book1996ndp}, \cite{book2008rl}.
	Originally introduced by \cite{1988td}, temporal-difference (TD) learning has become one of the most widely employed RL algorithms.
	 Another major breakthrough in RL was the development of a TD control algorithm, known as Q-learning \cite{thesis1989}, on which much of the contemporary artificial intelligence (see e.g., \cite{2015drl}) is built.	  
	Despite their popularity thanks to the simple updates they feature, theoretical analysis of RL (with function approximation) has proved challenging; see, for example, \cite{1994tdc},  \cite{ml1994qlearning}, \cite{1995tdd}, \cite{tac1997td}, \cite{book2008rl},
	  \cite{2008qlearning}. 
	  Moreover, non-asymptotic performance guarantees appeared only recently, and they still remain limited \cite{2015tdf}, \cite{2017error}, \cite{dalal2018td}, \cite{2018go}, \cite{bhandari2018td}, \cite{srikant2019td}, \cite{finite2019zou},  \cite{hu2019td}.
	
	Targeting a deeper understanding for the statistical efficiency of basic RL (e.g., TD and Q-learning) algorithms, the objective of this present paper is to derive non-asymptotic guarantees for a certain class of (biased) SA procedures. In particular, we first characterize a set of easy-to-check conditions on the nonlinear operators used in SA updates, and introduce a mild assumption on the stochastic noise sequence satisfied by a broad family of discrete-time stochastic processes. We prove a general convergence result based on a carefully constructed multi-step Lyapunov function, which relies on a number (as needed) of future SA updates to gain control over the gradient bias arising from instantaneous stochastic perturbations. For an introduction to Lyapunov theory, see e.g., \cite{lyapunov}, \cite{qinyuzhen}.
	We further develop non-asymptotic bounds on the mean-square error of the iterates.
	Finally, for direct comparison to past contributions, we specialize the results for general SA algorithms to both the TD-learning as well as the Q-learning with linear function approximation, from data gathered along a single trajectory of a Markov chain. 
	We thereby obtain finite-time error bounds for both TD and Q-learning algorithms using (non-)linear function approximators in the case of constant stepsizes, under the most general assumptions to date. 
	The merits of our bounds are that they apply to i) the unmodified TD as well as Q-learning algorithms (in sharp contrast, e.g., a projection step is required by \cite{dalal2018td}, \cite{bhandari2018td}, 
	\cite{finite2019zou}); 
	ii) nonlinear function approximators and Markov chains having general mixing rates 
	(bounds in \cite{bhandari2018td}, \cite{srikant2019td},  \cite{finite2019chen} were derived based on linear function approximation and geometric mixing); and, iii) Markov chains starting from any initial distribution as well as from the first iteration (meaning there is no need to wait until the Markov chain gets ``close'' to its unique stationary distribution as required by e.g., \cite{bhandari2018td}, \cite{srikant2019td}, \cite{finite2019chen}). 
	
	The remainder of this paper is structured as follows. Section~\ref{sec:nsa} begins with the basic background on SA and the formal problem formulation. Section \ref{sec:finite} presents the novel multi-step Lyapunov function followed by the main results on the non-asymptotic convergence guarantees for general SA algorithms. In Section \ref{sec:td}, we demonstrate the consequences of our main results for the problem of approximate reinforcement learning (i.e., TD$(0)$ and Q-learning using linear function approximators), and develop finite-time error bounds. Finally, the paper is concluded with research outlook in Section \ref{sec:conc}, while technical proofs of the main results
	are provided in the Appendix.


\textbf{Notation.} Throughout the paper, lower- (upper-) case letters denote deterministic (random) quantities of suitable dimensions clear from the context, e.g., $\theta$ ($\Theta$). Calligraphic letters stand for sets, e.g. $\mathcal{S}$, symbol $(\cdot)^\top$ represents transpose of matrices or vectors, and
$\|\theta\|_2$ (or simply, $\|\theta\|$) denotes the Euclidean norm of vector $\theta$. We reserve notation such as $c$, $c'$ etc. for some constants
that do not depend on any parameters of the considered MDPs, including the discount factor $\gamma$, or size of state
and action spaces, and so on.
	
\section{Background and problem set-up}
	\label{sec:nsa}
	
	Consider the following nonlinear recursion with a constant stepsize $\epsilon>0$, starting from $\Theta_0\in\mathbb{R}^d$
	\begin{equation}\label{eq:syst}
	{\Theta}_{k+1}={\Theta}_k+\epsilon f(\Theta_k, X_k),\quad k=0,1,2\ldots
	\end{equation}
	where $\Theta_{k} \in\mathbb{R}^d$ denotes the $k$-th iterate,
 $\{X_{k}\in\mathbb{R}^m \}_{k\in\mathbb{N}}$ is a stochastic noise sequence defined on a complete probability space,
 and $f:\mathbb{R}^d\times \mathbb{R}^m\to \mathbb{R}^d
 $ is a continuous function of  $(\theta,x)$. In the simplest setting, for example, $\{X_k\}_k$ is an independent and identically distributed (i.i.d.) random sequence of vectors, and $f(\Theta_k,X_k)$ is a conditionally unbiased estimate of the gradient $\wbar{f}(\Theta_k):= \mathbb{E}[f(\Theta_k,X_{k})|\mathcal{F}_k ]$. Here, $(\mathcal{F}_k)_{k\ge 0}$ is an increasing family of $\sigma$-fields, with $\Theta_0$ being $\mathcal{F}_0$-measurable, and $f(\theta,X_k)$ being $\mathcal{F}_k$-measurable. 
 	For an introduction to conditional expectation and $\sigma$-fields, see e.g., \cite{bookprob}.   Depending on whether $\mathcal{F}_0$ is a trivial $\sigma$-field, the initial guess $\Theta_0$ can be random or deterministic. With no loss of generality, the rest of this paper works with a deterministic $\Theta_0$.
In a more complicated setting pertaining to e.g., MDPs, $\{X_k\}_k$ is a Markov chain assumed to have a unique stationary distribution, and $f(\Theta_k,X_k)$ can be viewed as a biased estimate of some gradient $\bar{f}(\Theta_k)=		\lim_{k\to\infty} \mathbb{E}_{X_k}[f(\Theta_k,X_{k}) ]$.
	 		In both cases, we are prompted to assume that the following limit exists for each  $\theta\in\mathbb{R}^d $
	 		\begin{equation}
	 				\wbar{f}(\theta)=
	 				\lim_{k\to\infty} \mathbb{E}[f(\theta,X_{k}) ]
	 				\label{eq:limitexist}.
	 		\end{equation}
	 		
	 		Taking a dynamical systems viewpoint \cite{book2009sa}, the corresponding ordinary differential equation (ODE) for \eqref{eq:syst} is given by
	 		\begin{equation}
	 		\label{eq:cont}
	 		\dot{\theta}(t)=\wbar{f}(\theta(t)).
	 		\end{equation}
Assume that this ODE admits an equilibrium point $\theta^\ast$ at the origin, i.e., 
 $\wbar{f}(0)=0$. 
 This assumption is made without loss of generality, as one can always shift a nonzero equilibrium point to zero by appropriate centering $\theta\leftarrow \theta-\theta^\ast$. 
  Following the terminology in \cite{book2009sa}, 
  \cite{maei2009td},
   the recursion \eqref{eq:syst} is termed (nonlinear) stochastic approximation. Our goal here is to provide a non-asymptotic convergence analysis of the iterate sequence $\{\Theta_k\}_{k\in\mathbb{N}^+}$ generated by a recursion of the form \eqref{eq:syst} to the equilibrium point $\theta^\ast$ of its corresponding ODE \eqref{eq:cont}. 
 
 The motivating impetus for considering recursion \eqref{eq:syst} was to gain a deeper insight into the classical TD as well as Q-learning algorithms \cite{book2008rl} from discounted MDPs and reinforcement learning \cite{book2008rl}, \cite{book1996ndp}.
 It is a (biased) SA procedure for solving a fixed point equation defined by the so-called Bellman's operator \cite{tac1997td}. As a matter of fact, a large family of TD-based algorithms, including TD($0$), TD($\lambda$), and GTD, as well as stochastic gradient descent for nonlinear least-squares estimation can be described in this form (see e.g., \cite{2018go} for a detailed discussion).

		Certainly, convergence guarantees of SA procedures as in \eqref{eq:syst} would not be possible without imposing assumptions on the operators $f(\theta,x)$ and $\wbar{f}(\theta)$. In this work, motivated by the analysis of TD-learning and related algorithms in reinforcement learning, we consider a class of SA procedures that satisfy the following properties.
	\begin{assumption}\label{as:fxy}
		The function $f(\theta,x) $ satisfies the globally Lipschitz condition in $\theta$, uniformly in $x$, i.e., there exists a constant $L_1>0$ such that for all $\theta,\,\theta'\in\mathbb{R}^d$ and each $x\in\mathcal{X}$, it holds that 
		\begin{equation}\label{eq:lips}
		\left\| f(\theta,x)-f(\theta',x)\right\|\le L_1\|\theta-\theta'\|.
		\end{equation}   
Moreover, there exists a constant $L_2>0$ such that, for each $x\in\mathcal{X}$, it holds for all $\theta$
		\begin{equation}
		\label{eq:fbound}
		\|f(\theta,x)\|\le L_2(\|\theta\|+1)
		\end{equation}
		where $\mathcal{X}\subseteq\mathbb{R}^m $ denotes the living space of the stochastic process $\{X_{k} \}$. 
	\end{assumption}
	
	It is worth mentioning that \eqref{eq:fbound} is equivalent to assuming that $f(0,x)$ 
	satisfying \eqref{eq:lips} is uniformly bounded for all $x\in\mathcal{X}$. To see this, suppose $ \|f(0,x)\|\le \hat{f}$ holds for all $x\in\mathcal{X}$. From \eqref{eq:lips}, it follows that 
	$\|f(\theta,x)\|\le L_1\|\theta-\theta'\|+\|f(\theta',x)\|$, in which taking $\theta'=0$ confirms that $\|f(\theta,x)\|\le L_1\|\theta\|+\|f(0,x)\|\le L_1\|\theta\|+\hat{f}\le  \max(L_1,\hat{f})\cdot(\|\theta\|+1)$.
				In this regard, by defining $L:=\max \{L_1,L_2\}$, we will assume for simplicity that \eqref{eq:lips} and \eqref{eq:fbound} hold with the same constant $L$. 
		
	\begin{assumption}
		\label{as:lyap}
Consider the ODE \eqref{eq:cont}. There exists a twice differentiable function $W(\theta)$ that satisfies globally and uniformly the following conditions for all $\theta,\,\theta'\in\mathbb{R}^d$
		\begin{subequations}
			\label{eq:lyap}
			\begin{align}
			c_1\|\theta\|^2\le W(\theta) & \le c_2\|\theta\|^2\label{eq:lyap1} \\
			\left(\left.\frac{\partial W}{\partial \theta}\right|_{\theta}\right)^\top \wbar{f}(\theta)&\le - c_3L\|\theta\|^2 \label{eq:lyap2}\\
			\left\|\left. \frac{\partial W}{\partial \theta}\right|_{\theta}-\left. \frac{\partial W}{\partial \theta}\right|_{\theta'}\right\|&\le c_4\|\theta-\theta'\|\label{eq:lyap3}
			\end{align}    
		\end{subequations}
		for some constants $c_1,c_2,c_3,c_4>0$. 
	\end{assumption}

 Regarding these assumptions, two remarks come in order.
			\begin{remark}\label{rmk:func}
				Assumption \ref{as:fxy} is standard and widely adopted in the convergence analysis of SA algorithms; see e.g., \cite[Chapter 3]{book2009sa}, \cite{nips2011sa}, \cite{1994tdc}, \cite{ml1994qlearning}, and also \cite{tac1997td},
				  \cite{2018go}, \cite{srikant2019td} in the case of linear SA (i.e., $f(\theta,x)$ is linear in $\theta$). 
			\end{remark}		
			
 \begin{remark}
	By evaluating inequality \eqref{eq:lyap1} at $\theta=0$, one confirms that $W(\theta)>W(0)=0$ for all $\theta\ne 0$. Since $W(\theta)$ is twice differentiable, it implies that 
	$\frac{\partial W}{\partial \theta}|_{\theta=0}=0$.
	From \eqref{eq:lyap2}, it holds that both $\overline{f}(\theta)\ne 0$ and $\frac{\partial W}{\partial \theta}|_{\theta}\ne 0$ at any point $\theta\ne 0$. In words, Assumption \ref{as:lyap} states that the equilibrium point $\theta=0$ is unique, and globally, asymptotically stable for the ODE \eqref{eq:cont}. This also appeared in e.g.,  \cite[A5]{book2009sa} and \cite{nips2011sa} (strongly convex case).
	This is in the same spirit of
	requiring a Hurwitz matrix $\overline{A}$ (i.e., every eigenvalue has strictly negative real part) for the ODE
	$\dot{\theta}=\overline{A}\theta$ in linear SA by \cite[Theorem  2]{tac1997td}, 
	\cite{dalal2018td}, 
	\cite{srikant2019td}.
\end{remark}
			
				In addition to Assumptions \ref{as:fxy} and \ref{as:lyap}, to leverage the corresponding ODE to study convergence of SA procedures,  
				we make an assumption on the stochastic perturbation sequence $\{ X_k\}_{k\in\mathbb{N}}$.  
				
						\begin{assumption}\label{as:ergo}
 For each $\theta\in\mathbb{R}^d$, 
			 there exists a function $\sigma(T;T_0):\mathbb{N}^+\times \mathbb{N}^+\to \mathbb{R}^+$
			 monotonically decreasing to zero as either $T\to\infty$ or $T_0\to\infty$; i.e., $\lim_{T\to\infty}\sigma(T;T_0)=0$ for any fixed $T_0\in\mathbb{N}^+$, and $\lim_{T_0\to\infty}\sigma(T;T_0)=0$ for any fixed $T\in\mathbb{N}^+$,   such that
\begin{equation}\label{eq:ergo}
\left\|	\frac{1}{T}\!\sum_{k=T_0}^{T_0+T-1}\! \mathbb{E}\!\left[ f(\theta,X_{k})
\big|X_{0}
\right]-\wbar{f}(\theta)\right\|\le\sigma(T;T_0)L (\|\theta\|+1 )
							\end{equation}
							where the expectation $\mathbb{E}$ is taken over $\{X_{k} \}_{k=T_0}^{T_0+T-1}$. 
						\end{assumption}
			

 Simply put, Assumption \ref{as:ergo} requires that the bias of the `ergodic' average of any consecutive $T$ gradient estimates $\{ f(\theta,X_k)\}_{k=T_0}^{T_0+T-1}$ from their limit $\overline{f}(\theta)$ vanishes (at least) sublinearly in $T$. Indeed, this is fairly mild and more  general than those studied by e.g., \cite{nips2011sa}, \cite{2018go}, \cite{bhandari2018td}, \cite{srikant2019td}, each of which imposes requirements on every single gradient estimate $f(\theta,X_k)$.
In sharp contrast, our condition \eqref{eq:ergo} 
can allow for large instantaneous biased gradient estimates $f(\Theta_k,X_k)$ of $\wbar{f}(\Theta_k)$. Indeed, Assumption \ref{as:ergo} is satisfied by a broad family of discrete-time stochastic processes, including e.g., i.i.d. random vector sequences \cite{nips2011sa}, finite-state irreducible and aperiodic Markov chains \cite{2010mcmc}, and Ornstein-Uhlenbeck processes \cite{2019process}; whereas, existing works \cite{nips2011sa}, \cite{2018go}, \cite{bhandari2018td},  \cite{srikant2019td} focus solely on one type of those stochastic processes.


\section{Non-asymptotic bounds on the mean-square error}
\label{sec:finite}
			In this paper, we seek to develop novel tools for proving non-asymptotic bounds on the mean-square error of the iterates $\{\Theta_k\}_{k\ge 1}$ generated by a recursion of the form \eqref{eq:syst} (to the equilibrium point $\theta^\ast=0$).
			Before presenting the main results, we start off by introducing an instrumental result which is the key to our novel approach to controlling the possible bias in gradient estimates
			 of the SA procedure. Its proof is provided in Appendix \ref{pf:prop} of the supplementary material.
	
	\begin{proposition}\label{prop:exist}
		Under Assumptions \ref{as:fxy}
		 and \ref{as:ergo}, there exists a function $g'(k,T,\Theta_{k})$ such that the next relation holds for all $T\in\mathbb{N}^+$
		\begin{equation}
		\label{eq:tele}
		\Theta_{k+T}=\Theta_{k}+\epsilon T\wbar{f}(\Theta_{k})+ g'(k,T,\Theta_{k})
		\end{equation}
satisfying
		\begin{align}
		\left\|\mathbb{E}\big[g'(k,T,\Theta_k)
	\big|\Theta_{k} 
		\big]\!\right\|\le&\, \epsilon LT \beta_k(T,\epsilon) (\|\Theta_k\|+1)\label{eq:gfun}\\
		\beta_k(T,\epsilon) :=&\,
		\epsilon L T(1+\epsilon L)^{T-2}+\sigma(T;k)\label{eq:delt}
		\end{align}
		where the expectation is taken over $\{X_j \}_{j=k}^{k+T-1}$. 
	\end{proposition}
	
	Evidently, Proposition \ref{prop:exist} offers a bound on the average gradient bias over some $T>0$ iterations, which is indeed motivated by our Assumption \ref{as:ergo}.  
Based on the results in Proposition \ref{prop:exist}, we present the following theorem, which establishes a general convergence result that applies to any stochastic sequence $\{X_k\}_{k\in\mathbb{N}}$ satisfying Assumption \ref{as:ergo}.
		\begin{theorem}
			\label{th:exist}
			Under Assumptions \ref{as:fxy}---\ref{as:ergo} and for any $\delta>0$,
			there exist
			a function $W'(k, \Theta_k)$, and constants $(T^\ast\in\mathbb{N}^+,\epsilon_{\delta}) $ such that  $\sigma(T^\ast;k)\le \delta$ and the following inequalities are globally and uniformly satisfied for all $\epsilon\in (0,\,\epsilon_{\delta})$ and all $k\in\mathbb{N}$
			\begin{align}
			\label{eq:exis1}
			c_1' \|\Theta_k\|^2\le W'(k,\Theta_k)&\le c_2' \|\Theta_k\|^2+c_2''(\epsilon L)^2\\
			\mathbb{E}\big[W'(k+1,	\Theta_{k+1}
			)-W'(k,\Theta_k)
			\big|\Theta_{k}
			\big]&\le -\epsilon c_3' \|\Theta_k\|^2+c_4'\epsilon^2 + c_5' \sigma(T^\ast;k) \epsilon
			\label{eq:exis2}
			\end{align}
 where $c_1',c_2',c_3',c_2'',c_4',c_5'>0$ are constants dependent on $c_1\!\sim \! c_4$ of \eqref{eq:lyap} but independent of $\epsilon>0$.
		\end{theorem}
		
								Proof of Theorem \ref{th:exist} is relegated to Appendix \ref{pf:exist} of the supplementary material due to space limitations.  Our proof builds critically on the construction of 
								 function $W'(k, \Theta_k)$ from the Lyapunov function $W(\theta)$ of the ODE \eqref{eq:cont}. To be able to use the concentration bound in \eqref{eq:ergo},
				we are motivated to introduce a function candidate that necessarily looks ahead to a number of $T$ future iterates, with parameter $T\in\mathbb{N}^+$ to be designed such that the gradient bias 
				 can be made affordable, given by
				\begin{equation}\label{eq:candi}
				W'(k,\Theta_k)=\sum_{j=k}^{k+T-1}\! W(\Theta_j(k,\Theta_k))
				\end{equation}
				where, to make the dependence of $\Theta_{j\ge k}$ as a function of $\Theta_k$ explicit, we intentionally write $\Theta_j=\Theta_j(k,\Theta_k) $, understood as the iterate of the recursion \eqref{eq:syst} at time instant $j\ge k$, with an
				initial condition $\Theta_k$ at time $k$. The design parameter $T\ge 1$ allows us to exploit the monotonically decreasing function $\sigma(T;k)\to 0$ in \eqref{eq:gfun} to gain control over the gradient bias, therefore rendering the bounds \eqref{eq:exis1}--\eqref{eq:exis2} possible. When $\{X_k\}_{k}$ is e.g., an i.i.d. random sequence \cite{nips2011sa}, or a Markov chain that has approximately arrived at its steady state (i.e., after a certain mixing time of recursions) \cite{srikant2019td}, they have shown it suffices to choose $T=1$, that is $W'(\Theta_k)=W(\Theta_k)$ to validate \eqref{eq:exis1}--\eqref{eq:exis2}. For general Markov chains however, functions like $W'(\Theta_k)=W(\Theta_k)$ may fail to yield finite-time error bounds that hold for the entire sequence $\{\Theta_k\}_{k\ge 1}$. 
				In a nutshell, our novel way of constructing the Lyapunov function offers an effective tool for finite-time analysis of general SA algorithms driven by a broad family of (discrete-time) stochastic processes.

		
			We are now ready to study the drift of $W'(k,\Theta_k)$, which follows from Theorem \ref{th:exist}, and whose proof is provided in Appendix \ref{pf:recursive} of the supplementary material.
			\begin{lemma}
				\label{le:recursive}
				Under Assumptions \ref{as:fxy}---\ref{as:ergo}, the following holds true for all $\epsilon\in (0,\epsilon_{\delta})$ and all $k\in\mathbb{N}$
				\begin{equation}
				\label{eq:vdrift}
				\mathbb{E}\big[W'(k+1,\Theta_{k+1})
				\big]\le \left(1- \frac{c_3'\epsilon}{c_2'} \right) \mathbb{E}\big[W'(k,\Theta_k)
				\big]+ c_4''\epsilon^2 +c_5' \sigma(T^\ast;k) \epsilon
				\end{equation}
				where $c_4''>0$ is an appropriate constant independent of $\epsilon$, and $T^\ast\in\mathbb{N}^\ast$ is fixed in Theorem \ref{th:exist}.
			\end{lemma}
			
			\begin{theorem}
				\label{th:finite}
				Let $k_\epsilon:=\min\{k\in\mathbb{N}^+|\sigma(T^\ast;k)\le \epsilon \}$.  Under Assumptions \ref{as:fxy}---\ref{as:ergo}, and choosing any stepsize $\epsilon\in (0,\epsilon_{\delta})$, the following finite-time error bounds hold for all $k\in\mathbb{N}$
				\begin{equation}
				\label{eq:finite}
\mathbb{E}\big[\|\Theta_k\|^2\big]\le \frac{c_2'}{c_1'}\!\left(1- \frac{c_3'\epsilon}{c_2'} \right)^{\!k}\!\|\Theta_0\|^2+\frac{c_2'' L^2}{c_1'}\epsilon^2+ \frac{ c_6}{c_1'}\epsilon+\frac{ c_6}{c_1'}
\!\left(1- \frac{c_3'\epsilon}{c_2'} \right)^{\max\{k-k_\epsilon,0\}}\!\delta
				\end{equation}
				where $c_6>0$ is some constant; and $\delta>0$ is given in Theorem \ref{th:exist}.
			\end{theorem}
	
	Our bound in Theorem \ref{th:finite} reveals a two-phase convergence behavior of biased SA schemes obeying Assumptions \ref{as:fxy}---\ref{as:ergo}. At the beginning, since the gradient bias characterized in terms of $\sigma(T^\ast;k)$ is sizable; precisely, $\sigma(T^\ast;k)\gg \epsilon$, the last term in \eqref{eq:finite} is basically a constant $(c_6/c_1')\delta$, that does not shrink as $\epsilon\to0$. That is,
	the biased SA procedure converges linearly fast only to a constant-size [size-$(c_6/c_1')\delta$] neighborhood of the equilibrium point $\theta^\ast=0$ of the associated ODE. On the other hand, as $k$ increases, the gradient bias $\sigma(T^\ast;k)$ decreases. Specifically, when $k\ge k_\epsilon$, or $\sigma(T^\ast;k)\le \epsilon$, the size of this neighborhood can be controlled by $\epsilon$. Under this condition, we establish that the SA recursion converges linearly fast to an $\epsilon$-neighborhood of the equilibrium point $\theta^\ast=0$, which can be made arbitrarily small by setting $\epsilon>0$ small enough.
	
Proof of Theorem \ref{th:finite} is postponed to Appendix \ref{pf:finite} of the supplemental document. At this point, some observations are worth making. 
\begin{remark}
Existing non-asymptotic results have focused on either linear SA algorithms (e.g., \cite{liu2015td}, \cite{2018go}, \cite{srikant2019td}, \cite{bhandari2018td}, \cite{icml2019td}), or nonlinear SA under i.i.d. noise (e.g., \cite{siam2009sa}, \cite{nips2011sa}). In sharp contrast, the finite-time error bound in Theorem \ref{th:finite} is applicable to a class of nonlinear SA procedures under a broad family of discrete-time stochastic perturbation processes. 
\end{remark}

\begin{remark}
When the general recursion \eqref{eq:syst} is specialized to linear SA driven by Markovian noise $\{X_k\}_{k\in\mathbb{N}}$, i.e., $f(\Theta_k,X_k)=A(X_k)\Theta_k + b(X_k)$, 
our established bound in \eqref{eq:finite} improves upon the state-of-the-art in \cite[Theorem  7]{srikant2019td} by a constant $\sim(1- {c_3'\epsilon}/{c_2'} )^{\tau}$, where $\tau\gg 1$ is the mixing time of the Markov chain $\{X_k\}$. In fact, the bound in \cite[Theorem  7]{srikant2019td} becomes applicable only after a mixing time of updates (i.e., for $k\ge \tau$) till the Markov chain gets sufficiently `close' to its stationary distribution; yet, in sharp contrast, our bound \eqref{eq:finite} is effective from the first iteration for Markov chains starting with any initial distribution. 
Furthermore, our stead-state value (the last term of \eqref{eq:finite}) scales only with the stepsize $\epsilon>0$ (which has removed the independence on $\tau$ from the bound in \cite{srikant2019td}), and it vanishes as $\epsilon\to 0$.	
\end{remark}

 Evidently, with the bound in \eqref{eq:finite}, one can easily estimate the number of data samples (e.g., the length of a Markov chain trajectory) required for the mean-square error to be of the same order as its steady-state value.

	\section{Applications to approximate reinforcement learning}\label{sec:td}
	We now turn to the consequences of our general results for the problem of reinforcement learning with (non-)linear function approximation (a.k.a., approximate reinforcement learning), in particular the TD- and Q-learning.
	Toward this objective, we begin by providing a brief introduction to discounted MDPs and basic reinforcement learning algorithms; interested readers can refer to standard sources (e.g., \cite{book2008rl}, \cite{book1996ndp}) for more background.

	\subsection{Background and set-up}
Consider an MDP, defined by the 
quintuple $(\mathcal{S},
\mathcal{U},
\mathcal{P},R,\gamma)$,
where $\mathcal{S}$ is a finite set of possible states (a.k.a. state space), $\mathcal{U}$ is a finite set of possible actions (a.k.a. action space), 
$\mathcal{P}:=\{P^u\in\mathbb{R}^{|\mathcal{S}|\times|\mathcal{S}|}|u\in\mathcal{U} \}$ is a collection of probability transition matrices, indexed by actions $u$,
$R(s,u):\mathcal{S}\times \mathcal{U}\to \mathbb{R}$ is a reward received upon executing action $u$ while in state $s$, 
and $\gamma\in [0,1)$ is the discount factor. The results along with theoretical analysis developed in this paper may be generalized to deal with infinite and compact state and/or action spaces, but we restrict ourselves to finite spaces here for an ease of exposition.

An agent selects actions to interact with the
MDP (the environment) by operating a policy.
Specifically, at each time step $k\in\mathbb{N}$, the agent first observes the state $S_k=s\in\mathcal{S}$ of the environment, and takes an action $U_k=u\in\mathcal{U}$ by
following a deterministic policy $\pi:\mathcal{S}\to\mathcal{U}$, or a stochastic one $U_k\sim \pi(\cdot|S_k)$, where $\pi(\cdot|s)$ is a probability distribution function supported on $\mathcal{U}$. The environment then moves to the next state $S_{k+1}=s'\in\mathcal{S}$ with probability $P_{ss'}^u={\Pr}(S_{k+1}=s'|S_k=s,U_k=u)$, associated with which an instantaneous reward $R_k:=R(S_k,U_k) $ is revealed to the agent. 
This procedure generates a trajectory of states, actions, and rewards, namely, $S_0,U_0,R_0,S_1,\ldots,S_T,U_T,R_T,S_{T+1},\ldots$ over $\mathcal{S}\times\mathcal{U}\times\mathbb{R}$.

For a given policy $\pi$, the value function $V^{\pi}(s):\mathcal{S}\to\mathbb{R}$ measures the expected discounted reward obtained by starting the MDP in a given state, and then following the policy $\pi$ to take actions in all subsequent iterations. 
More precisely, we define
\begin{equation}
V^{\pi}(s)=\mathbb{E}\!\left[\sum_{k=0}^{\infty} \gamma^k R(S_k,U_{k}) \Big|S_0=s\right]
\end{equation}
where
the conditional expectation $\mathbb{E}$ is taken over the entire trajectory of observations conditioned on the initial state $S_0=s$.

Let $R^{\pi}(s):\mathcal{S}\to\mathbb{R}$ with $R^{\pi}(s):=\sum_{s'\in\mathcal{S}}\sum_{u\in\mathcal{U}} \pi(u|s)P_{ss'}^u R(s,u)$, and $P^{\pi}(s,s'):\mathcal{S}\times\mathcal{S}\to\mathbb{R}$ with $P^{\pi}(s,s'):=\sum_{u\in\mathcal{U}}\pi(u|s)P^{u}_{ss'}$.
Postulating a canonical ordering on the elements of $\mathcal{S}$, we can view $V^{\pi}$ and $R^{\pi}$ as vectors in $\mathbb{R}^{|\mathcal{S}|}$, and $P^{\pi}$ as a matrix in $R^{|\mathcal{S}|\times|\mathcal{S}|}$. 
It is well known that such a function $V^{\pi}$ exists given uniformly bounded rewards $\{R(s,u)\le \bar{r}\}_{s,u}$, and it is the unique solution to the next Bellman's equation \cite{tac1997td,maei2009td}
\begin{equation}\label{eq:vs}
V^{\pi}(s)=R^{\pi}(s)+\gamma\sum_{s'\in\mathcal{S}} P^{\pi}(s,s')   V^{\pi}(s'),\quad \forall s\in\mathcal{S}.
\end{equation}
Evidently, if all transition probabilities $\{P^{\pi}(s,s')  \}_{s,s'}$ were known, vector $V^\pi$ 
can be found by solving a system of $|\mathcal{S}|$ linear equations dictated by \eqref{eq:vs}. 
But in practice, either $\{P^{\pi}(s,s')  \}_{s,s'}$ are unknown or $|\mathcal{S}|$ is huge or even infinity, it is impossible to evaluate exactly $V^\pi$. 
Here, the goal is to efficiently estimate the value function $V^\pi(s)$ by observing a single trace of the Markov chain $\{S_0,
S_1,S_2,\ldots\}$ and the associated rewards, which is to be dealt with in subsection \ref{subsec:td}.  Since the policy $\pi$ we consider in this paper will be assumed fixed, we shall drop the dependence on $\pi$ in our notation for brevity. 

Likewise, we can define for control purpose the so-called action-value function (a.k.a., Q-function), which measures the quality of a given policy by
the expected sum of discounted instantaneous rewards, conditioned on starting in a given state-action pair, and following the policy $\pi$ to take subsequent actions; i.e.,      
\begin{equation}
\label{eq:qfunc}
Q(s,u)=\mathbb{E}\!\left[\sum_{k=0}^{\infty} \gamma^k R(S_k,U_k)
\Big|S_0=s,U_0=u\right],\quad {\rm where~} U_k\sim\pi(\cdot|S_k)~{\rm for~all~} k\in\mathbb{N}^+.
\end{equation}
Naturally, we would like to choose the policy $\pi$ such that the values of the Q-function are optimized. Indeed, it has been established that the Q-function associated with the optimal policy $\pi^\ast$, yielding the optimal Q-function denoted by $Q^\ast$, satisfies the following Bellman equation \cite{book1996ndp,ml1994qlearning,qlearning}
\begin{equation}
\label{eq:qbell}
Q^\ast(s,u)=\mathbb{E}[R(s,u)]+\gamma \mathbb{E}\!\left[ \max_{u'\in\mathcal{U}}\,
Q^\ast(s',u')\Big|s,u \right].
\end{equation}
after assuming a canonical ordering on the elements of $\mathcal{S}\times \mathcal{U}$, the table $Q$ can be treated as a matrix in $\mathbb{R}^{|\mathcal{S}|\times|\mathcal{U}|}$.
Once 
$\{Q^\ast(s,u)\}_{s,u}$
becomes available, an optimal policy $\pi^\ast$ can be recovered by setting
$\pi^\ast(s) \in \arg\max_{u\in\mathcal{U}}Q^\ast (s,u)$ for all $s\in\mathcal{S}$, without any knowledge about the transition probabilities.

Again, in the learning context of interest, the transition probabilities $\{P^u_{ss'} \}_{s,u,s'}$ are unknown and the dimensions $|\mathcal{S}|$ and/or $|\mathcal{U}|$ can be huge or even infinity in practice, so it is not possible to exactly evaluate the Bellman equation \eqref{eq:qbell}. As one of the most popular solutions for finding the optimal policy, Q-learning \cite{thesis1989} iteratively updates the estimate $Q_k$ of $Q^\ast$ using a single trajectory of samples $\{(S_k,U_k,S_{k+1})\}$ generated by following the policy $\pi$, 
according to the recursion
\begin{equation}
\label{eq:qupdate}
Q_{k+1}(S_k,U_k)=Q_k(S_k,U_k)+\epsilon_k\!\left[R(S_k,U_k)+\gamma\max_{u'\in\mathcal{U}} \, Q_k(S_{k+1},u')-Q_k(S_k,U_k)
\right]
\end{equation}
where $\{0<\epsilon_k<1 \}$ is a sequence of stepsizes to be chosen by the user. Under standard conditions on the stepsizes, the sequence $\{ Q_k\}$ converges to $Q^\ast$ almost surely as long as every state-action pair $(s,u)\in\mathcal{S}\times\mathcal{U}$ is visited infinitely often; see, for instance, \cite{qlearning,ml1994qlearning,book1996ndp,book2008rl}. 

However, it is well known that for many important problems of practical interest, the computational requirements of exact function estimation are overwhelming, mainly because of a large number of states and actions (i.e., Bellman's ``curse of dimensionality'')  \cite{book1996ndp}. Instead, a popular approach in the literature has been to leverage low-dimensional parametric approximations of the value function, or the Q-function.
Although contemporary nonlinear approximators such as  neural networks \cite{2015drl}, \cite{tsp2019wgc} could lead to more powerful approximations, the simplicity of reinforcement learning with linear function approximation \cite{book2008rl} allows us to analyze them in detail.

\subsection{Temporal-difference learning with linear function approximation}\label{subsec:td}

In this section, we target a deeper understanding of the dynamics of TD-learning algorithms with linear function approximation. Toward this objective, we assume a linear function approximator for the true value function $V(s)$, given by
\begin{equation}
V(s)\approx V^{\theta}(s)=\phi^\top\! (s)\theta
\end{equation}
where $\theta\in\mathbb{R}^{d}$ is a parameter vector to be learned, typically having $d\ll  |\mathcal{S}|$; and $\phi(s)\in \mathbb{R}^d$ is a fixed feature vector dictated by preselected basis functions $\phi_1,\phi_2,\ldots,\phi_d:\mathcal{S}\to\mathbb{R}$, that act on state $s$ as follows
\begin{equation}
\phi(s):=[\phi_1(s)~\phi_2(s)~\cdots~\phi_n(s) ]^\top\in\mathbb{R}^d,\quad \forall s\in\mathcal{S}.
\end{equation}

Without loss of generality, we assume that all feature vectors are $\|\phi(s)\|\le 1$, $\forall s\in\mathcal{S}$ \cite{maei2009td}.
For future reference, we stack up all feature vectors $\{\phi(s) \}_{s\in\mathcal{S}}$ to form the matrix 
\begin{equation*}
\Phi:=\left[\begin{array}{c}
\phi^\top(s_1)\\
\phi^\top(s_2)\\
\vdots\\
\phi^\top(s_{|\mathcal{S}|})
\end{array}
\right]\in\mathbb{R}^{|\mathcal{S}|\times d}.
\end{equation*}
We also make a standard assumption that
$\Phi$ has full column rank, i.e., any redundant or irrelevant feature vectors have been removed  \cite{tac1997td}

Consider the classical TD-learning for estimating the unknown parameter vector $\theta$ \cite{1988td}. Starting with some $\Theta_0$, the simplest yet widely used TD variant, which is also known as TD($0$), updates the estimate $\Theta_k$ of $\theta$ according to
\begin{equation}
\label{eq:td0}
\Theta_{k+1}=\Theta_k -\epsilon \left[\phi^\top\! (S_k)\Theta_k - R(S_k,U_{k}) -\gamma \phi^\top\!(S_{k+1})\Theta_k \right]\!\phi(S_k)
\end{equation}
where $\epsilon \in (0,1)$ is a constant stepsize. We are interested in developing finite-time error bounds for \eqref{eq:td0}, where the observed tuples $\{(S_k,R(S_k,U_{k}),S_{k+1})\}_{k\in\mathbb{N}}$ are gathered from a single trajectory of the Markov chain $\{S_k\}_{k\in\mathbb{N}}$.
To apply our results developed for general SA algorithms 
to TD($0$) learning here, we just need to verify our working assumptions in Section \ref{sec:nsa}.

Let $\mu \in [0,1)^{|\mathcal{S}|}$ also denote the stationary distribution of the Markov chain, i.e., $\mu=\mu P$,
and also $D\in\mathbb{R}^{|\mathcal{S}|\times |\mathcal{S}|}$ be a diagonal matrix holding entries of $\mu$ on its main diagonal. 
Upon introducing $A:=-\Phi D(\Phi-\gamma P \Phi)^\top $ and $b:=\Phi D R$, it has been shown in \cite[Theorem  2]{tac1997td} that the TD($0$) algorithm with diminishing stepsizes obeying standard conditions, tracks the ODE
\begin{equation}
\dot{\theta}= A \theta + b\label{eq:lode}
\end{equation}  
as well as converges to its unique equilibrium point $\theta^\ast=-A^{-1}b$. 
By substituting $ \Theta= \tilde{\Theta}+\theta^\ast$ into \eqref{eq:td0}, we deduce that
the centered recursion 
\begin{align*}
\tilde{ \Theta}_{k+1}=\tilde{ \Theta}_k + \epsilon \phi(S_k) \! \left[\gamma \phi^\top\!(S_{k+1})\tilde{ \Theta}_k-\phi^\top\! (S_k)\tilde{ \Theta}_k +R(S_k,U_{k}) +\gamma \phi^\top\!(S_{k+1}) \theta^\ast-\phi^\top\! (S_k) \theta^\ast  \right]
\end{align*} 
tracks the ODE $\dot{\tilde{\theta}}=A\tilde{\theta}$, which has a unique equilibrium point at the origin $\tilde{\theta}=0$.  
If we define
\begin{align}
X_k:=&\, [S_k~S_{k+1}]^\top\nonumber\\
f(\tilde{\Theta}_k, X_k):= &\, \phi(S_k)\!\left[\gamma \phi(S_{k+1})-\phi(S_k)\right]^\top\! \tilde{\Theta}_k + R(S_k,U_{k}) \phi(S_k)\label{eq:0xk} \\
& +\phi(S_k)\!\left[\gamma \phi(S_{k+1})-\phi(S_k)\right]^\top\! \theta^\ast \label{eq:0fxy}
\end{align}
it now becomes clear that the TD($0$) algorithm
is a special case of the stochastic recursion \eqref{eq:syst}.  

As has also been made clear in our discussion above, the following limit exists for each $\tilde{\theta}\in\mathbb{R}^d$
\begin{equation}
\wbar{f}(\tilde{\theta})= \lim_{k\to \infty} \mathbb{E}[ f(\tilde{\theta}, X_k)]= A \tilde{\theta},\quad{\rm with}\quad  \wbar{f}(0)=0.
\end{equation}

Next, let us turn to verify Assumptions \ref{as:fxy}---\ref{as:ergo}.
\newline

\noindent\emph{\textbf{Verifying Assumption \ref{as:fxy}.}} For any $\tilde{\theta},\,\tilde{\theta}'\in\mathbb{R}^d$, 
consider $X=x=[s~s']^\top$ with $s,s'\in \mathcal{S}$. We have that
\begin{align}
\big\| f(\tilde{\theta},X)- f(\tilde{\theta}',X)\big\|& =\left\|\phi(s)\!\left[\gamma \phi(s')-\phi(s)\right]^\top\! \big(\tilde{\theta}-\tilde{\theta}'\big)
\right\|\nonumber\\
&\le \left\|\gamma \phi(s)\phi^\top(s')-\phi(s)\phi^\top(s)\right\| \big\|\tilde{\theta}-\tilde{\theta}'\big\|\nonumber\\
&\le \big[ \gamma \|\phi(s)\| \|\phi(s')\| +\|\phi(s)\| \|\phi(s')\| 
\big] \big\|\tilde{\theta}-\tilde{\theta}'\big\|\nonumber\\
&\le 2\big\|\tilde{\theta}-\tilde{\theta}'\big\|\label{eq:0flip}
\end{align} 
for all possible $s,s'\in \mathcal{S}$, 
where the last inequality follows from the assumption that $\| \phi(s)\|\le 1$ for all $s\in\mathcal{S}$ and $0\le \gamma <1$. Clearly, inequality \eqref{eq:lips} holds with constant $L_1=2$. 

Likewise, for each $\tilde{\theta}$, it follows for all $s,s'\in\mathcal{S}$ that
\begin{align}
\big\|f(\tilde{\theta},X)\big\|&=\left\|\phi(s)\left[\gamma \phi(s')-\phi(s)\right]^\top\! \tilde{\theta} +R(s,u) \phi(s)
+\phi(s)\!\left[\gamma \phi(s')-\phi(s)\right]^\top \!\theta^\ast
\right\|\nonumber\\
&\le \left\|\gamma\phi(s) \phi(s')-\phi(s)\phi(s)\right\| \|\tilde{\theta}\| + |R(s,u)| \| \phi(s)\|
+\left\|\phi(s)\!\left[\gamma \phi(s')-\phi(s)\right]^\top\right\| \|\theta^\ast
\|\nonumber\\
&\le 2\|\tilde{\theta}\| + \bar{r} + 2\|\theta^\ast\|\label{eq:0fbound}\\
&\le \max(2, \bar{r} + 2\|\theta^\ast\|) (\|\tilde{\theta}\|+1)\label{eq:0fbound1}
\end{align}
where \eqref{eq:0fbound} is due to the uniformly bounded rewards $\{|R(s,u)|\le \bar{r}\}$, verifying the inequality \eqref{eq:fbound}. 
By combining \eqref{eq:0flip} and \eqref{eq:0fbound1}, Assumption \ref{as:fxy} holds with constant $L:=\max(2, \bar{r}+2\|\theta^\ast\|)$. 
\newline

\noindent\emph{\textbf{Verifying Assumption \ref{as:lyap}.}} 
Consider now the centered ODE $\dot{\tilde{\theta}}=\wbar{f}(\tilde{\theta})=A\tilde{\theta}$. It has been shown in \cite{tac1997td} that $A=-\Phi D(\Phi-\gamma P \Phi)^\top$ is negative definite (but not symmetric), in the sense that $ \tilde{\theta}^\top A\tilde{\theta}<0$ for all $\tilde{\theta}\ne 0$; that is, $A$ is Hurwitz and full rank. So, the equilibrium point $\tilde{\theta}=0$ is globally, asymptotically stable for this ODE. Based on linear system theory, one can construct a Lyapunov function candidate $V(\tilde{\theta})$ as follows 
\begin{equation}
V(\tilde{\theta})= \tilde{\theta}^\top P \tilde{\theta}\label{eq:0lyap}
\end{equation}
for some symmetric, positive definite matrix $P\in\mathbb{R}^{d\times d}$ to be determined.

It is clear that $\lambda_{\min}(P) \|\tilde{\theta}\|^2\le V(\tilde{\theta})\le \lambda_{\max}(P) \|\tilde{\theta}\|^2$
for all $\tilde{\theta}\in\mathbb{R}^d$, where $\lambda_{\max}(P)\ge \lambda_{\min}(P)>0$ denote the smallest and largest eigenvalues of $P$, respectively. Hence, inequality \eqref{eq:lyap1} holds with constants $c_1=\lambda_{\min}(P)$ and $c_2=\lambda_{\max}(P)$. 

Regarding \eqref{eq:lyap2}, we have that
\begin{equation*}
\left(\frac{\partial V}{\partial \tilde{\theta}}\Big|_{\tilde{\theta}}\right)^\top\! \wbar{f}(\tilde{\theta})=2\tilde{\theta}^\top P A\tilde{\theta}=\tilde{\theta}^\top\! \left(A^\top P + PA\right)\tilde{\theta}. 
\end{equation*}
Consider the following continuous Lyapunov equation with Hurwitz $A$ 
\begin{equation}
A^\top P + P A=-Q
\end{equation}
which is well known to have a unique symmetric, positive definite solution $P$ given any symmetric, positive definite matrix $Q$ (e.g., \cite{book1996ndp}).
So, by choosing any $Q=Q^\top\succ 0$, we verified that \eqref{eq:lyap2} holds with constant $c_3=\lambda_{\min}(Q)/L>0$ as well.

As far as \eqref{eq:lyap3} is concerned, observing that $ \big\|\frac{\partial V}{\partial \tilde{\theta}}\big|_{\tilde{\theta}}\big\|=2\|P\tilde{\theta}\|\le 2\lambda_{\max}(P) \|\tilde{\theta}\| $, it suffices to take $c_4=2\lambda_{\max}(P)$. Summarizing these three cases, we have shown that Assumption \ref{as:lyap} is met by the TD($0$) algorithm.
\newline

\noindent\emph{\textbf{Verifying Assumption \ref{as:ergo}.}} 
This following property holds for any finite, irreducible, and aperiodic Markov chain \cite[Theorem 4.9]{mcbook}. 
\begin{lemma} 
	\label{le:mix}
	For any finite-state, irreducible, and aperiodic Markov chain, there are constants $c_0>0$ and $\rho\in (0,1)$ such that
	\begin{equation}\label{eq:mix}
	\max_{x\in\mathcal{X}} \,d_{TV}\big({\rm Pr}(X_k\in \cdot |X_0=x),\mu \big)\le c_0\rho^k,\quad \forall k\in\mathbb{N}
	\end{equation}
	where $\mu$ is the steady-state distribution of $\{X_k\}_{k\in\mathbb{N}}$, and  $d_{TV}(\nu,\mu)$ denotes the total variation distance between probability measures $\mu$ and $\nu$, defined as follows
	\begin{equation}
	d_{TV}(\nu,\mu):=\frac{1}{2}\sum_{x\in\mathcal{X}}\big|\nu(x)-\mu(x) \big|.
	\end{equation}	
\end{lemma}

Considering $X_{T_0}=x_{T_0}=[s_0~s_0']^\top$ with any $s_0,s_0'\in\mathcal{S}$, it holds for each $\tilde{\theta}\in\mathbb{R}^d$ that
\begin{align}
& \!\left\|	\frac{1}{T}\!\sum_{k=T_0+1}^{T_0+T} \mathbb{E}\big[  f(\tilde{\theta},X_{k})\big|X_{0}=[s_0~s_0']\big]-\wbar{f}(\tilde{\theta})\right\|\nonumber\\
=\, & \bigg\|	\frac{1}{T}\!\sum_{k=T_0+1}^{T_0+T} \mathbb{E}\Big[  
\phi(S_k)\!\left[\gamma \phi(S_{k+1})-\phi(S_k)\right]^\top\! \tilde{\theta} + R(S_k,U_{k}) \phi(S_k)\nonumber\\
&
+\phi(S_k)\!\left[\gamma \phi(S_{k+1})-\phi(S_k)\right]^\top\! \theta^\ast\Big| S_{0}=s_0,S_{1}=s_0' \Big]-\mathbb{E}_{\mu}[f(\tilde{\theta},X) ]\bigg\|\nonumber\\
=\, & \bigg\|	\frac{1}{T}\!\sum_{k=T_0+1}^{T_0+T}
\sum_{s\in\mathcal{S}}  \big[{\rm Pr}(S_{k}=s|S_{0}=s_0,S_{1}=s_0')-\mu(s) \big]\Big[  \phi(s)\!\left(\gamma P(s,s') \phi(s')-\phi(s)\right)^\top\! \tilde{\theta} \label{eq:0ergo2}\\
&+ r(s) \phi(s)
+\phi(s)\!\left(\gamma P(s,s')\phi(s')-\phi(s)\right)^\top\! \theta^\ast
\Big] \bigg\|\nonumber\\
\le \, &\max_{s,s'} \Big\|
\phi(s)\!\left(\gamma P(s,s') \phi(s')-\phi(s)\right)^\top\! \tilde{\theta} + r(s) \phi(s)
+\phi(s)\!\left(\gamma P(s,s')\phi(s')-\phi(s)\right)^\top\! \theta^\ast
\Big\|
\nonumber\\
&\,\times 	\frac{1}{T}\!\sum_{k=T_0+1}^{T_0+T}
\sum_{s\in\mathcal{S}}  \big|{\rm Pr}(S_{k}=s|S_{1}=s_0')-\mu(s) \big| \label{eq:0ergo100}\\
\le\, & L(\|\tilde{\theta}\|+1) \times \frac{1}{T}\!\sum_{k=T_0+1}^{T_0+T} 2 c_0 \eta^{k-1}\label{eq:0mix}\\
\le \,& \sigma(T;T_0)
L(\|\tilde{\theta}\|+1)
\end{align}
where $\mathbb{E}_{\pi}[f(\tilde{\theta},X) ]$ takes expectation over $X=[S~S']^\top$ with respect to the stationary distribution $\mu$ of the Markov chain $\{S_k\}$; \eqref{eq:0ergo2} leverages the transition probabilities $\{P_{ss'} \}$; \eqref{eq:0mix} uses the inequality \eqref{eq:0fbound1} and the fact that finite-state, irreducible, and aperiodic Markov chains mix at a geometric rate $\eta\in(0,1)$ from any $s_0'\in\mathcal{S}$ (see Lemma \ref{le:mix}). 
Now, we can conclude that Assumption \ref{as:ergo} holds with $L:=\max(1+\gamma,\bar{r})$ and
function  $\sigma(T;T_0):=\frac{2c_0 \eta^{T_0}}{(1-\eta) T} 
 $, which indeed monotonically decreases to $0$ as $T\to\infty$ or $T_0\to \infty$. 

\begin{remark}\label{rmk:mix}
	It is worth remarking that although geometric mixing is used here for ease of presentation,
	it is clear from \eqref{eq:0ergo100} that Assumption \ref{as:ergo} is satisfied by Markov chains with general mixing rates as long as $\sum_{k=T_0+1}^{T_0+T} \sum_{s\in\mathcal{S}}  \big|{\rm Pr}(S_{k}=s|S_{1}=s_0')-\mu(s) \big| $ grows sub-linearly in $T$; that is,
	\begin{equation}
	\label{eq:general}
	\lim_{T\to\infty} \frac{1}{T}\sum_{k=T_0+1}^{T_0+T} \sum_{s\in\mathcal{S}}  \big|{\rm Pr}(S_{k}=s|S_{T_0+1}=s_0')-\mu(s) \big| =0,\quad \forall s_0'\in\mathcal{S},\, T_0\in\mathbb{N}.
	\end{equation}
\end{remark}

In a nutshell, the finite-time error bound in Theorem \ref{th:finite} applies to TD($0$) learning algorithms in a generally mixing Markov chain setting, provided standard assumptions---that the reward function is uniformly bounded as $\max_{s,s'}|R(s,u)|\le \bar{r}$, the feature matrix $\Phi$ is full column rank, and all feature vectors $\{\phi(s)\}$ are bounded too---hold true. 

\begin{remark}
	Although the basic TD(0) learning algorithm is dealt with here, it is worth remarking that other variants of TD-learning such as TD($\lambda$), and GTD can also be cast in the form of \eqref{eq:syst}; see related treatments in e.g., \cite{srikant2019td}. Hence, our novel multistep Lyapunov approach as well as the associated non-asymptotic convergence analysis is readily applicable to general TD-learning algorithms. 
\end{remark}


\subsection{Q-learning with linear function approximation}
\label{subsec:qlearning}

In this section, we provide an improved non-asymptotic analysis for Q-learning using linear function approximators. In this direction, we assume that the Q-function is parameterized by a linear function as follows
\begin{equation}
\label{eq:qapp}
Q(s,u)\approx Q^{\theta}(s,u)=\psi^\top(s,a)\theta
\end{equation}
where we have kept the notation $\theta\in\mathbb{R}^d$ for the parameter vector to be learned (as in the previous subsection), typically of size $d\ll |\mathcal{S}|\times|\mathcal{U}|$, the number of state-action pairs; and the feature vector $\psi(s,a)\in\mathbb{R}^d$ stacks up $d$ features produced by pre-selected basis functions $\{\psi_{\ell}(s,u):\mathcal{S}\times\mathcal{U}\to \mathbb{R}\}_{\ell=1}^d$. Similar to TD-learning with linear function approximation, here we can also introduce the feature matrix, given by
\begin{equation*}
\Psi:=\left[\begin{array}{c}
\psi^\top(s_1,u_1)\\
\psi^\top(s_1,u_2)\\
\vdots\\
\psi^\top(s_{|\mathcal{S}|},u_{|\mathcal{U}|})
\end{array}
\right]\in\mathbb{R}^{|\mathcal{S}||\mathcal{U}|\times d}
\end{equation*}
which is assumed to have full column rank (that is, linearly independent columns) and satisfy $\|\psi(s,u)\|\le 1$ for all state-action pairs $(s,u)\in\mathcal{S}\times\mathcal{U}$. 

The well-known approximate Q-learning algorithm updates the parameter vector $\Theta$, according to the recursion (e.g., \cite{book1996ndp,book2008rl})
\begin{equation}
\label{eq:qupd}
\Theta_{k+1}=\Theta_k +\epsilon \psi(S_k,U_k)\!\left[
R(S_k,U_k)+\gamma\max_{u\in\mathcal{U}} \psi^\top(S_{k+1},u) \Theta_k-\psi^\top(S_k,U_k)\Theta_k
\right]
\end{equation}
with some constant stepsize $\epsilon\in (0,1)$. 
The objective here is to establish finite-time error guarantees for \eqref{eq:qupd}, when the observed data samples
$\{(S_k,U_k,R(S_k,U_k),S_{k+1},U_{k+1}) \}_{k\in\mathbb{N}}$
are collected along a single path of the Markov chain $\{S_k\}_{k\in\mathbb{N}}$ by following a deterministic policy $\pi$. 
Considering 
\begin{equation}
F(\theta,X_k)\!=\psi(S_k,U_k)\!\left[
R(S_k,U_k)\!+\gamma\max_{u\in\mathcal{U}} \psi^\top(S_{k+1},u) \theta\!-\psi^\top(S_k,U_k)\theta
\right],\, {\rm where}\, X_k:=(S_k,U_k,S_{k+1})
\end{equation}  
it becomes obvious that \eqref{eq:qupd} has the form of \eqref{eq:syst}.
To check whether our non-asymptotic error guarantees established for nonlinear SA procedures in Section \ref{sec:finite} can be applied to Q-learning with linear function approximation or not, it, again, suffices to show that Assumptions \ref{as:fxy}---\ref{as:ergo} are satisfied by the Q-learning updates \eqref{eq:qupd}. 

In general, Q-learning with function approximation can diverge. This is mainly because Q-learning implements off-policy \footnote{On-policy methods estimate the value of a policy while using it for control (namely, taking actions); while in off-policy methods, the policy used to generate behavior, called the behavior/sampling policy, may be independent of the policy that is evaluated and improved, called the target/estimation policy \cite{book2008rl}.
} sampling to collect the data, which renders the expected Q-learning update possibly an expansive mapping \cite{1995stable}. 
Under appropriate regularity conditions on the sampling policy, 
asymptotic convergence of Q-learning with linear function approximation was established in \cite{2008qlearning}, and finite-time analysis was recently given in \cite{finite2019zou,finite2019chen}. 
In the following, we also impose some regularity conditions on 
the sampling policy $\pi$. 
\begin{assumption}[\cite{finite2019chen}]
	\label{as:policy} Suppose that the Markov chain $\{ S_k\}_{k\in\mathbb{N}}$ induced by policy $\pi$ is irreducible and aperiodic, whose unique stationary distribution is denoted by $\mu$. Assume that  
	the equation $\bar{F}(\theta):=\mathbb{E}_{\mu}[F(\theta,X)]=0$ has a unique solution $\theta^\ast$, and 
	the next inequality holds for all $\theta\in\mathbb{R}^d$ 
	\begin{equation}
	\gamma^2\mathbb{E}_{\mu}\! \left[\max_{u'\in\mathcal{U}} \big(\psi^\top\!(s',u')\,\theta \big)^2 \right]-\mathbb{E}_{\mu}\!\left[\big(\psi^\top\!(s,u)\,\theta \big)^2
	\right]\le -c\|\theta\|^2,\quad{\rm where~} u\sim \pi(\cdot|s)
	\label{eq:sampling}
	\end{equation}
	for some constant $0<c<1$.
\end{assumption}


Now, let us turn to verify Assumptions \ref{as:fxy}---\ref{as:ergo}.
Toward this end, we start by introducing $\tilde{ \theta}:=\theta-\theta^\ast$ and $X:=(S,U,S')$. It then follows that 
\begin{equation}\label{eq:fdef}
f(\tilde{\theta}):=F(\tilde{\theta}+\theta^\ast,X)=\psi(S,U)\!\left[
R(S,U)+\gamma\max_{u\in\mathcal{U}} \psi^\top(S',u) \theta-\psi^\top(S,U)\theta
\right] 
\end{equation}
It is then evident that $\bar{f}(\tilde{\theta}):=\mathbb{E}_{\mu}[{F}(\tilde{\theta}+\theta^\ast,X)]=0$ has a unique solution $\tilde{\theta}^\ast=0$.
Now, we can rewrite \eqref{eq:qupd} as follows
\begin{equation}
\tilde{\Theta}_{k+1}=\tilde{\Theta}_k +\epsilon {f}(\tilde{\Theta}_k,X_k). \label{eq:qcentered}
\end{equation}
\newline

\noindent\emph{\textbf{Verifying Assumption \ref{as:fxy}.}}
For any $\tilde{\theta}_1,\,\tilde{\theta}_2$ and $x=(s,u,s')$, we have that
\begin{align}
\left\|{f}(\tilde{\theta}_1,x)-\bar{f}(\tilde{\theta}_2,x)\right\|&=\Big\|\psi(s,u)\Big[R(s,u)+\gamma\max_{u_1\in\mathcal{U}}\psi^\top(s',u_1)\big(\tilde{ \theta}_1+{ \theta}^\ast \big)-\psi^\top(s,u)\big(\tilde{ \theta}_1+{ \theta}^\ast \big)
\Big]\nonumber\\
&~\;\,\quad -
\psi(s,u)\Big[R(s,u)+\gamma\max_{u_2\in\mathcal{U}}\psi^\top(s',u_2)\big(\tilde{ \theta}_2+{ \theta}^\ast \big)-\psi^\top(s,u)\big(\tilde{ \theta}_2+{ \theta}^\ast \big)
\Big]
\Big\|\nonumber\\
&=\Big\|\gamma\psi(s,u)\Big[\max_{u_1\in\mathcal{U}}\psi^\top(s',u_1)\big(\tilde{ \theta}_1+{ \theta}^\ast \big)-\max_{u_2\in\mathcal{U}}\psi^\top(s',u_2)\big(\tilde{ \theta}_2+{ \theta}^\ast \big)
\Big]\nonumber\\
&~\;\,\quad +\psi(s,u)\psi^\top(s,u)\big(\tilde{\theta}_1-\tilde{\theta}_2 \big)\Big\|\nonumber\\
&\le \gamma\Big|\max_{u_1\in\mathcal{U}}\psi^\top(s',u_1)\big(\tilde{ \theta}_1+{ \theta}^\ast \big)-\max_{u_2\in\mathcal{U}}\psi^\top(s',u_2)\big(\tilde{ \theta}_2+{ \theta}^\ast \big)\Big|+\big\|\tilde{\theta}_1-\tilde{\theta}_2\big\|\label{eq:vas11}
\end{align}
where the last inequality follows from $\|\psi(s,u)\|\le 1$ for all $(s,u)\in\mathcal{S}\times\mathcal{U}$.

On one hand, suppose that $u_1^\ast\in \max_{u_1\in\mathcal{U}}\psi^\top(s',u_1)\big(\tilde{ \theta}_1+{ \theta}^\ast \big)$, then
\begin{align}
\max_{u_1\in\mathcal{U}}\psi^\top\!(s',u_1)\big(\tilde{ \theta}_1+{ \theta}^\ast \big)-\max_{u_2\in\mathcal{U}}\psi^\top\!(s',u_2)\big(\tilde{ \theta}_2+{ \theta}^\ast \big)&=\psi^\top\!(s',u_1^\ast)\big(\tilde{ \theta}_1+{ \theta}^\ast \big)-\max_{u_2\in\mathcal{U}}\psi^\top\!(s',u_2)\big(\tilde{ \theta}_2+{ \theta}^\ast \big)\nonumber\\
&\le \psi^\top\!(s',u_1^\ast)\big(\tilde{ \theta}_1+{ \theta}^\ast \big)-\psi^\top\!(s',u_1^\ast)\big(\tilde{ \theta}_2+{ \theta}^\ast \big)\nonumber\\
&=\psi^\top\!(s',u_1^\ast)\big(\tilde{ \theta}_1-\tilde{ \theta}_2 \big)\nonumber\\
&\le \big\|\tilde{\theta}_1-\tilde{\theta}_2\big\|\label{eq:onehand}
\end{align}
due again to $\|\psi(s',u_1^\ast)\|\le 1$.
On the other hand, if we let $u_2^\ast\in \max_{u_2\in\mathcal{U}}\psi^\top(s',u_2)\big(\tilde{ \theta}_2+{ \theta}^\ast \big)$, it follows similarly that
\begin{align}
\max_{u_1\in\mathcal{U}}\psi^\top\!(s',u_1)\big(\tilde{ \theta}_1+{ \theta}^\ast \big)-\max_{u_2\in\mathcal{U}}\psi^\top\!(s',u_2)\big(\tilde{ \theta}_2+{ \theta}^\ast \big)&=\max_{u_1\in\mathcal{U}}\psi^\top\!(s',u_1)\big(\tilde{ \theta}_1+{ \theta}^\ast \big)-\psi^\top\!(s',u_2^\ast)\big(\tilde{ \theta}_2+{ \theta}^\ast \big)\nonumber\\
&\ge \psi^\top\!(s',u_2^\ast)\big(\tilde{ \theta}_1+{ \theta}^\ast \big)-\psi^\top\!(s',u_2^\ast)\big(\tilde{ \theta}_2+{ \theta}^\ast \big)\nonumber
\\
&=\psi^\top\!(s',u_2^\ast)\big(\tilde{ \theta}_1-\tilde{ \theta}_2 \big)\nonumber\\
&\ge -\big\|\tilde{\theta}_1-\tilde{\theta}_2\big\|.
\label{eq:otherhand}
\end{align}

Combining \eqref{eq:onehand} and \eqref{eq:otherhand} yields
\begin{equation}
\Big|\max_{u_1\in\mathcal{U}}\psi^\top(s',u_1)\big(\tilde{ \theta}_1+{ \theta}^\ast \big)-\max_{u_2\in\mathcal{U}}\psi^\top(s',u_2)\big(\tilde{ \theta}_2+{ \theta}^\ast \big)\Big|\le \big\|\tilde{\theta}_1-\tilde{\theta}_2\big\|
\end{equation}
which, in conjunction with \eqref{eq:vas11}, proves that
\begin{equation}
\left\|{f}(\tilde{\theta}_1,x)-{f}(\tilde{\theta}_2,x)\right\|\le (\gamma+1)\big\|\tilde{\theta}_1-\tilde{\theta}_2\big\|.\label{eq:vas12}
\end{equation}
In the meanwhile, it is easy to see that
\begin{align}
\big\|{f}(\tilde{ \theta},x)\big\|&=\Big\|\psi(s,u)\Big[R(s,u)+\gamma\max_{u_1\in\mathcal{U}}\psi^\top(s',u_1)\big(\tilde{ \theta}+{ \theta}^\ast \big)-\psi^\top(s,u)\big(\tilde{ \theta}+{ \theta}^\ast \big)
\Big]\Big\|\nonumber\\
&\le |R(s,u)|+\gamma \|\psi(s',u_1^\ast)\| \big\|\tilde{ \theta}+{ \theta}^\ast \big\|+\|\psi(s,u)\|\big\|\tilde{ \theta}+{ \theta}^\ast\big \|\nonumber\\
&\le \bar{r}+ (\gamma+1)\big(\|\tilde{ \theta}\|+\|{\theta}^\ast\|\big)\nonumber\\
&=(\gamma+1)\|\tilde{ \theta}\|+ \big[\bar{r}+ (\gamma+1)\|{\theta}^\ast\|\big]\label{eq:vas13}
\end{align}
where we have used the fact that $|R(s,u)|\le \bar{r}$ for all $(s,u)\in\mathcal{S}\times \mathcal{U}$.
With \eqref{eq:vas12} and \eqref{eq:vas13}, we have proved that Assumption \ref{as:fxy} is met with $L:=\max\{\gamma+1, \,\bar{r}+ (\gamma+1)\|{\theta}^\ast\| \}$.
\newline

\noindent\emph{\textbf{Verifying Assumption \ref{as:lyap}.}}
The ODE associated with the (centered) Q-learning update \eqref{eq:qcentered} is
\begin{equation}
\dot{\tilde{ \theta}}=\bar{f}(\tilde{ \theta})=\mathbb{E}_{\mu}\Big\{\psi(s,u)\Big[R(s,u)+\gamma \max_{u'\in\mathcal{U}}\psi^\top\!(s',u')\big(\tilde{ \theta}+\theta^\ast\big)-\psi^\top\!(s,u)\big(\tilde{ \theta}+\theta^\ast\big)
\Big]
\Big\}
\end{equation}
for which we consider the Lyapunov candidate function $W(\tilde{ \theta})=\|\tilde{ \theta}\|^2/2$. Evidently, it follows that $W(\tilde{ \theta})\ge 0$ for all $\tilde{ \theta}\ne 0$, so \eqref{eq:lyap1} holds with $c_1=c_2=1/2$. Secondly, using $\bar{f}(\tilde{ \theta}^\ast)=0$, we have that
\begin{align}
&\Big(\frac{\partial W(\tilde{ \theta})}{\partial \tilde{ \theta}}\Big)^\top\!\bar{f}(\tilde{ \theta})\nonumber\\
&=\Big(\frac{\partial W(\tilde{ \theta})}{\partial \tilde{ \theta}}\Big)^\top\!\Big[\bar{f}(\tilde{ \theta})-\bar{f}(\tilde{ \theta}^\ast)\Big]\nonumber\\
&={\tilde{ \theta}}^\top \mathbb{E}_{\mu}\Big\{\psi(s,u)\Big[R(s,u)+\gamma \max_{u_1\in\mathcal{U}}\psi^\top\!(s',u_1)\big(\tilde{ \theta}+\theta^\ast\big)-\psi^\top\!(s,u)\big(\tilde{ \theta}+\theta^\ast\big)
\Big]\nonumber\\
&\qquad\qquad  -\psi(s,u)\Big[ R(s,u)+\gamma \max_{u_2\in\mathcal{U}} \psi^\top(s',u_2)\theta^\ast-\psi^\top(s,u)\theta^\ast\Big]
\Big\}\nonumber\\
&=\gamma \mathbb{E}_{\mu}\Big\{\tilde{\theta}^\top\psi(s,u)\Big[ \max_{u_1\in\mathcal{U}}\psi^\top\!(s',u_1)\big(\tilde{ \theta}+\theta^\ast\big)-\max_{u_2\in\mathcal{U}} \psi^\top(s',u_2)\theta^\ast\Big]\Big\} -\mathbb{E}_{\mu} \Big[\psi^\top\!(s,u)\tilde{\theta}\Big]^2\nonumber\\
&\le \gamma\sqrt{ \mathbb{E}_{\mu}\!\Big[\psi^\top\!(s,u)\tilde{\theta}\Big]^2}\sqrt{ \mathbb{E}_{\mu}\!\Big[ \max_{u_1\in\mathcal{U}}\psi^\top\!(s',u_1)\big(\tilde{ \theta}+\theta^\ast\big)\!-\!\max_{u_2\in\mathcal{U}} \psi^\top\!(s',u_2)\theta^\ast\Big]^2}\!-\mathbb{E}_{\mu}\! \Big[\psi^\top\!(s,u)\tilde{\theta}\Big]^2\label{eq:qas2eq2}\\
&= \sqrt{ \mathbb{E}_{\mu}\Big[\psi^\top\!(s,u)\tilde{\theta}\Big]^2 }\Bigg\{\gamma \sqrt{ \mathbb{E}_{\mu}\!\Big[ \max_{u_1\in\mathcal{U}}\psi^\top\!(s',u_1)\big(\tilde{ \theta}+\theta^\ast\big)\!-\!\max_{u_2\in\mathcal{U}} \psi^\top(s',u_2)\theta^\ast\Big]^2}- \sqrt{ \mathbb{E}_{\mu}\Big[\psi^\top\!(s,u)\tilde{\theta}\Big]^2}
\Bigg\}\nonumber
\\
&\le \sqrt{ \mathbb{E}_{\mu}\Big[\psi^\top\!(s,u)\tilde{\theta}\Big]^2 }\Bigg\{\gamma \sqrt{ \mathbb{E}_{\mu}\!\max_{u'\in\mathcal{U}}\! \left[\psi^\top\!(s',u')\tilde{ \theta}\right]^2}- \sqrt{ \mathbb{E}_{\mu}\Big[\psi^\top\!(s,u)\tilde{\theta}\Big]^2}\Bigg\}\label{eq:qas2eq31}\\
&= \frac{\sqrt{ \mathbb{E}_{\mu}\Big[\psi^\top\!(s,u)\tilde{\theta}\Big]^2 } \left\{\gamma^2\mathbb{E}_{\mu}\Big[\max_{u'\in\mathcal{U}} \!\big(\psi^\top\!(s',u')\tilde{ \theta}\big)^2\Big] -\mathbb{E}_{\mu}\Big[\psi^\top\!(s,u)\tilde{\theta}\Big]^2 \right\} }{\gamma \sqrt{ \mathbb{E}_{\mu}\max_{u'\in\mathcal{U}} \!\Big[\psi^\top\!(s',u')\tilde{ \theta}\Big]^2 }+\sqrt{ \mathbb{E}_{\mu}\Big[\psi^\top\!(s,u)\tilde{\theta}\Big]^2}}\label{eq:qas2eq32}\\
&\le \frac{-c\big\|\tilde{ \theta}\big\|^2 }
{ \sqrt{\gamma^2 \mathbb{E}_{\mu}\Big[\max_{u'\in\mathcal{U}} \!\big(\psi^\top\!(s',u')\tilde{ \theta}\big)^2\Big] \Big/ \mathbb{E}_{\mu}\Big[\psi^\top\!(s,u)\tilde{\theta}\Big]^2}+1}\label{eq:qas2eq4}\\
&\le \frac{-c\big\|\tilde{ \theta}\big\|^2 }
{ 2-c}
\end{align} 
which suggests that \eqref{eq:lyap2} holds with $c_3:=c/[(2-c)L]$.


On the other hand, it follows for any $\tilde{\theta},\,\tilde{\theta}'$ that
\begin{align*}
\left\|\frac{\partial W}{\partial \tilde{\theta}}\big|_{\tilde{\theta}}
-\frac{\partial W}{\partial \tilde{\theta}}\big|_{\tilde{\theta}'}
\right\|=\big\|\tilde{\theta}-\tilde{\theta}'\big\|
\end{align*}
validating \eqref{eq:lyap3} with $c_4=1$.
\newline

\noindent\emph{\textbf{Verifying Assumption \ref{as:ergo}.}} 
Let $P_{ss'}^u$ be the transition probability of the Markov chain $\{S_k\}_{k\in\mathbb{N}}$ from state $s$ to $s'$ after taking action $u$; and let $p^{(n)}_{ss'}$ be the $n$-step transition probability from state $s$ to $s'$ following policy $\pi$. 
Define then $X_k:=(S_k, U_k, S_{k+1})$, and it can be verified that $\{X_k\}_{k\in\mathbb{N}}$ is a Markov chain with state space $\mathcal{X}:=\{x=(s,u,s'):s\in\mathcal{S},\,\pi(u|s)>0,P_{ss'}^u>0 \}\subseteq \mathcal{S}\times \mathcal{U}\times \mathcal{S}$.
Next, we show that $\{X_k\}$ is aperiodic and irreducible.

Toward that end, consider two arbitrary states $x_i=(s_i,u_i,s_i'),\,x_j=(s_j,u_j,s_j')\in\mathcal{X}$. Since $\{S_k\}_k$ is irreducible, there exists an integer $n>0$ such that $p^{(n)}_{s_i',s_j}>0$. Using the definition of $\{X_k\}_k$, it follows that
\begin{align}\label{eq:n1tp}
p^{(n+1)}_{x_i,x_j}
=p^{(n)}_{s_i's_j}\pi(u_j|s_j) P^{u_j}_{s_js_j'}>0
\end{align}
which corroborates that the Markov chain $\{X_k\}_k$ is irreducible (e.g., \cite[Chapter 1.3]{mcbook}). 

To prove that $\{X_k\}_k$ is aperiodic, we assume, for the sake of contradiction, that $\{X_k\}_k$ is periodic with period $d\ge 2$. As $\{X_k\}_k$ has been shown irreducible, it follows readily that every state in $\mathcal{X}$ has the same period of $d$. Hence, for each state $x=(s,u,s')\in\mathcal{X}$, it holds that
$p^{(n+1)}_{x,x}=0
$ for all integers $n+1>0$ not divisible by $d$. Further, we deduce for any positive integer $(n+1)$
not divisible by $d$ that
\begin{align}
p^{(n+1)}_{s's'}=\sum_{s\in\mathcal{S}} p^{(n)}_{s's}p^{(1)}_{ss'}&=\sum_{s\in\mathcal{S}}p_{s's}^{(n)}\sum_{u\in\mathcal{U}}\pi(u|s)P^{u}_{ss'}
=\sum_{s\in\mathcal{S}}\sum_{u\in\mathcal{U}}p^{(n+1)}_{xx}=0\label{eq:pspsp}
\end{align}
where the last two equalities arise from \eqref{eq:n1tp} and the periodicity assumption of $\{X_k\}_k$, respectively. It becomes evident from \eqref{eq:pspsp} that $\{S_k\}_k$ is periodic too, and its period is at least $d$. This clearly contradicts with the assumption that $\{S_k\}_k$ is aperiodic. Therefore, we conclude that the Markov chain $\{X_k\}_k$ is irreducible and aperiodic provided that $\{S_k\}_k$ is irreducible and aperiodic. 

Now, consider two arbitrary states $x_{0}=(s_{0},u_{0},s_{0}'),\,x=(s,u,s')\in\mathcal{X}$. It follows that
\begin{align}
\bigg\|\frac{1}{T}\!\sum_{k=T_0+1}^{T_0+T}\!\mathbb{E}&\big[f(\tilde{ \theta},X_k)\big|X_{0}=x_{0}\big]-\bar{f}(\tilde{ \theta})
\bigg\|
=\bigg\|\frac{1}{T}\!\sum_{k=T_0+1}^{T_0+T}\sum_{x\in\mathcal{X}}\big[p_{x_{0}x}^{k}-\mu(x)\big]f(\tilde{ \theta},x)\bigg\|
\label{eq:arb0}\\
&=\bigg\|\frac{1}{T}\!\sum_{k=T_0+1}^{T_0+T} \sum_{s,s'\in\mathcal{S}}~\sum_{u\in\mathcal{U}}~\left[p_{s_{0}' s}^{k}\pi(u|s) P_{ss'}^{u}-\mu(x)
\right]\nonumber\\
&\quad \times  \psi(s,u)
\Big[R(s,u)+\gamma\max_{u'\in\mathcal{U}}\psi^\top\!(s',u')\big(\tilde{ \theta}+{ \theta}^\ast \big) -\psi^\top\!(s,u)\big(\tilde{ \theta}+{ \theta}^\ast \big)
\Big]
\bigg\|\label{eq:arb1}\\
&\le \max_{(s,u,s')\in\mathcal{X}}\bigg\|\psi(s,u)\Big[R(s,u)+\gamma\max_{u'\in\mathcal{U}}\psi^\top\!(s',u')(\tilde{ \theta}+\theta^\ast)-\psi^\top\!(s,u)\big(\tilde{ \theta}+{ \theta}^\ast \big)
\Big]
\bigg\|\nonumber\\
&\quad \times \frac{1}{T}\!\sum_{k=T_0+1}^{T_0+T} ~~\sum_{x=(s,u,s')\in\mathcal{X}}
\left|p_{s_{0}' s}^{k}\pi(u|s) P_{ss'}^{u}-\mu(x)
\right|\nonumber\\
&\le L\big(\|\tilde{ \theta}\|+1\big) \times\frac{1}{T}\sum_{k=T_0+1}^{T_0+T} 2c_0 \eta^{k-1}\label{eq:arb3}\\
&\le 
\sigma(T;T_0)
L\big(\|\tilde{ \theta}\|+1\big)\nonumber
\end{align} 
where \eqref{eq:arb0} is due to the definition that $\bar{f}(\tilde{ \theta})=\mathbb{E}_{X\sim \mu}[f(\tilde{ \theta},X)]=\sum_{x\in\mathcal{X}} \mu(x)f(\tilde{ \theta},x)$; equality \eqref{eq:arb1} uses \eqref{eq:fdef} and \eqref{eq:n1tp}; and, \eqref{eq:arb3} arises from the geometric mixing property of irreducible, aperiodic Markov chain $\{X_k\}_k$ as well as \eqref{eq:vas13}.

We have proved that Assumptions \ref{as:fxy}---\ref{as:ergo} are satisfied by Q-learning with linear function approximation, provided that certain conditions on the sampling policy and function approximators hold. Hence, the established finite-time error bound in Theorem \ref{th:finite} holds for Q-learning algorithms with linear function approximation.

\section{Comparison to past work}
Since the seminal work \cite{1988td}, there has been a large body of contributions on TD-learning in diverse settings. Here we shall focus on the subset of results that has established error bounds for discounted MDPs,
which are most relevant for direct comparison to our results.
General asymptotic results on the convergence of TD-learning algorithms with 
(non-)linear 
function approximation were given in \cite{1994tdc}, \cite{tac1997td}, 
\cite{maei2009td}; and those of Q-learning (with function approximation) in \cite{qlearning}, \cite{ml1994qlearning}, \cite{nips1998qlearning}, 
 \cite{2008qlearning}. Examples of divergence 
 with (non-)linear approximation  
were provided by \cite{1995tdd}, \cite{tac1997td}. 
Connections between TD-learning and SA were drawn in \cite{1994tdc}, \cite{ml1994qlearning}, and \cite{book1996ndp}. 

On the other hand, non-asymptotic guarantees for
RL algorithms
appeared only recently and remain still limited; see \cite{2015tdf}, \cite{2017error},
	\cite{liu2015td}, \cite{2017gtd}, 
	\cite{zhang2018finite}, 
	\cite{icml2019td}, \cite{2019wainwright}, \cite{finite2019zou}, \cite{finite2019chen}. Finite-time analysis of TD($0$) was initially investigated by \cite{2015tdf}, which, however, was pointed out to contain serious errors in the proofs by \cite{2017error} and thus not included in our comparison. 
Finite-time performance of other TD-based algorithms such as 
GTD was studied by \cite{
	liu2015td,2017gtd}, and \cite{icml2019td}. 
Refined finite-time error bounds of TD($0$) were provided by \cite{dalal2018td}, \cite{2018go}, but their results apply only to i.i.d. data samples, which, however, are difficult to acquire in practice. 
Dealing with the more realistic yet challenging Markov chain observation model, finite-time analysis 
  on the mean-square error 
was recently studied by \cite{bhandari2018td}, \cite{srikant2019td}, \cite{finite2019zou}, \cite{finite2019chen}, and \cite{2019wainwright}.
Nevertheless, their bounds were derived for TD- and/or Q-learning using linear function approximators, and were based on strong geometric mixing conditions. 
In addition, \cite{bhandari2018td} and \cite{finite2019chen} require including a projection step into the standard updates, whereas the bounds reported by \cite{srikant2019td} and \cite{finite2019chen} become applicable only after a certain mixing time of TD updates, that is, after the Markov chain gets sufficiently ``close'' to its stationary distribution. 
On the other hand, drawing connections between TD learning and Markov jump linear systems, exact convergence behaviors of the first- and second-order moments of a family of TD learning algorithms to their steady-state values were characterized using classical control theory in \cite{hu2019td}.  

In contrast, 
our bound in Theorem \ref{th:finite} not only applies to
TD- and Q-learning with linear function approximation, but also using a certain class of nonlinear function approximators compliant with Assumption \ref{as:fxy}. More importantly, our non-asymptotic guarantees hold for the unmodified TD- and Q-learning (i.e., without any projection steps), as well as for Markov chains under general mixing conditions as specified in \eqref{eq:general} and from any initial distribution.

\section{Conclusions}\label{sec:conc}

In this paper, we 
provided a non-asymptotic analysis for
a class of biased SA algorithms driven by a broad family of stochastic perturbations, which include as special cases e.g., i.i.d. random sequences of vectors and ergodic Markov chains. Taking a dynamical systems viewpoint, our approach has been to design a novel multistep Lyapunov function that involves future iterates to control the gradient bias. We proved a general convergence result based on this multistep Lyapunov function, and developed non-asymptotic bounds on the mean-square error of the iterate generated by the SA procedure to the equilibrium point of the associated ODE. Subsequently, we illustrated this general result by applying it to obtain finite-time error bounds for the unmodified TD- and Q-learning with linear function approximation, where
 data are gathered along a single trajectory of a Markov chain. Our bounds hold for Markov chains with general mixing rates and any initial distribution, as well as from the first iteration. Although the focus here has been on biased SA procedures with constant stepsizes, our non-asymptotic results can be extended to accommodate time-varying stepsizes as well. Our future work will also aim at generalizing this novel analysis to SARSA and distributed RL algorithms.

\bibliographystyle{IEEEtran}
\bibliography{learning}

%
%
%

\appendix


\section{Proof of Proposition \ref{prop:exist}}
\label{pf:prop}

We start off the proof by introducing the following auxiliary function
\begin{equation}
\label{eq:step1}
g(k,T,\Theta_{k}):=	\Theta_{k+T}- \Theta_{k}-\epsilon\sum_{j=k}^{k+T-1} f(\Theta_{k},X_{j} ),\quad \forall  T\ge 1
\end{equation}
which is evidently well defined under our working Assumptions \ref{as:fxy} and \ref{as:ergo}. Regarding the function $g(k,T,\Theta_{k})$ above, we present the following useful bound, whose proof details are, however, postponed to Appendix \ref{pf:le1} for readability.

\begin{lemma}
	\label{le:gbound}
	For any given $\Theta_k\in\mathbb{R}^d$, the function $g(k,T,\Theta_k)$ satisfies for all $k\ge 0$
	\begin{equation}\label{eq:gbound}
	\left	\|g(k,T,\Theta_k)\right\|\le \epsilon^2L^2 T^2(1+\epsilon L)^{T-2}(\|\Theta_k\|+1),\quad \forall T\ge 1.
	\end{equation}
\end{lemma}

On the other hand, note from \eqref{eq:tele} that
\begin{equation}
\label{eq:gprimedef}
g'(k,T,\Theta_{k})=\Theta_{k+T}-\Theta_{k}-\epsilon T\wbar{f}(\Theta_{k})
\end{equation}
which, in conjunction with \eqref{eq:step1}, suggests that we can write
\begin{align}\label{eq:gprimefunc}
g'(k,T,\Theta_k)&=
g(k,T,\Theta_k)+\epsilon\! \sum_{j=k}^{k+T-1} f(\Theta_k,X_{j})-\epsilon T\wbar{f}(\Theta_k)\nonumber\\
&=g(k,T,\Theta_k)+\epsilon\! \sum_{j=k}^{k+T-1}\!\left( f(\Theta_k,X_{j})-\wbar{f}(\Theta_k) \right).
\end{align}

Considering any given ${\Theta}_k$, if one
 takes expectation of both sides of \eqref{eq:gprimefunc}, 
 we obtain that
\begin{align}
\mathbb{E}\!\left[g'(k,T,\Theta_k)
 \big|{\Theta}_k
\right]
=\,&\mathbb{E}\!\left[ g(k,T,\Theta_k) 
\right]+\epsilon\,\mathbb{E}\! \left[\sum_{j=k}^{k+T-1}\!\left( f(\Theta_k,X_{j})-\wbar{f}(\Theta_k) \right) 
\Big|{\Theta}_k
\right]\nonumber\\
=\,& \mathbb{E}\!\left[ g(k,T,\Theta_k) 
\big|{\Theta}_k
\right]+\epsilon T\left(\frac{1}{T}\sum_{j=k}^{k+T-1} \mathbb{E}\!\left[  f(\Theta_k,X_{j})-\wbar{f}(\Theta_k)
\big|{\Theta}_k
\right]
\right)\nonumber\\
\le \,& \epsilon L T\!\left[\epsilon L T\!\left(1+\epsilon L\right)^{T-2} 
+\sigma(T;k)\right] (\|\Theta_k\|+1)\label{eq:gp4}
\end{align}
where the last inequality follows from
Lemma \ref{le:gbound} as well as the property of the averaged operator $\wbar{f}$ in \eqref{eq:ergo} under our working Assumption \ref{as:ergo}.
This concludes the proof.

\section{Proof of Theorem \ref{th:exist}}
\label{pf:exist}

We prove this theorem by carefully constructing function for $W'(k, \Theta_k)$ from $W (\Theta_k)$ (recall under our working assumption \ref{as:lyap} that $W(\Theta_k)$ exists and satisfies properties \eqref{eq:last1}---\eqref{eq:lyap3}).
Toward this objective, let us start with the following candidate
\begin{equation}\label{eq:vprimedef}
W'(k,\Theta_k)=\sum_{j=k}^{k+T-1} W(\Theta_j(k,\Theta_k))
\end{equation}
where, to make the dependence of $\Theta_{j\ge k}$ on $\Theta_k$ explicit, we maintain the notation $\Theta_j=\Theta_j(k,\Theta_k) $, which is understood as the state of the recursion \eqref{eq:syst} at time instant $j\ge k$, with an
initial condition $\Theta_k$ at time instant $k$. 

In the following, we will show that there exists and also determine a value for the parameter $T\in\mathbb{N}^+$ such that the inequalities \eqref{eq:exis1} and \eqref{eq:exis2} are satisfied. 

For ease of exposition, we start by proving the second inequality \eqref{eq:exis2}.
To this end, observe from the definition of $W'(k,\Theta_k)$ in \eqref{eq:vprimedef} that
\begin{align}
W'(k+1,\Theta_k+\epsilon f(\Theta_k,X_k))-W'(k,\Theta_k)
&=\sum_{j=k+1}^{k+T} W(\Theta_j(k,\Theta_k))-\sum_{j=k}^{k+T-1} W(\Theta_j(k,\Theta_k))\nonumber\\
&=W(\Theta_{k+T}(k,\Theta_k))-W(\Theta_k(k,\Theta_k))\nonumber\\
&=W(\Theta_{k+T}(k,\Theta_k))-W(\Theta_k)
\label{eq:vdiff}
\end{align}
where the last equality is due to the fact that $\Theta_k(k,\Theta_k)=\Theta_k$.

To upper bound the term in \eqref{eq:vdiff}, 
we will focus on bound the first term $W(\Theta_{k+T}(k,\Theta_k))$. Recall from \eqref{eq:tele} that
\begin{align*}
\Theta_{k+T}(k,\Theta_k)=\Theta_k+\epsilon T \wbar{f}(\Theta_k) +g'(k,T,\Theta_k)
\end{align*}
based on which we can find the second-order Taylor expansion of $W(\Theta_{k+T}(k,\Theta_k))$ (which is twice differentiable under Assumption \ref{as:lyap})
around $\Theta_k$, as follows
\begin{align}
W(\Theta_{k+T}(k,\Theta_k))&=W(\Theta_k)+\left(\left.\frac{\partial W}{\partial \theta}\right|_{\Theta_k}\right)^\top \left[\epsilon T \wbar{f}(\Theta_k) +g'(k,T,\Theta_k)\right]\nonumber\\
&\quad + \left[\epsilon T \wbar{f}(\Theta_k) +g'(k,T,\Theta_k)\right]^\top \nabla^2 W(\Theta_k')  \left[\epsilon T \wbar{f}(\Theta_k) +g'(k,T,\Theta_k)\right]\label{eq:taylor}
\end{align}
where we have employed the so-called mean-value theorem, suggesting that \eqref{eq:taylor} holds with $\Theta_k':=\Theta_k+\eta \big[\epsilon T \wbar{f}(\Theta_k) +g'(k,T,\Theta_k) \big]$ for some constant $\eta\in [0,1]$. 

Next, we will pursue an upper bound for each individual term on the right hand side of \eqref{eq:taylor} by conditioning on the $\sigma$-field $\mathcal{F}_k$. Again, 
invoking \eqref{eq:lyap2}, 
we have that
\begin{align}
\mathbb{E}\!\left[ \epsilon T \Big(\left.\frac{\partial W}{\partial \theta}\right|_{\Theta_k}\Big)^\top \wbar{f}(\Theta_k)\Big|\Theta_{k}  \right] \le -c_3\epsilon  LT \|\Theta_k\|^2\label{eq:term1}.
\end{align}

One can further verify the following bounds
\begin{align}
\mathbb{E}\!\left[	\left(\!\left.\frac{\partial W}{\partial \theta}\right|_{\Theta_k}\right)^\top \!g'(k,T,\Theta_k)\Big|\Theta_{k} \right]&=	\left(\left.\frac{\partial W}{\partial \theta}\right|_{\Theta_k}\right)^\top \! \mathbb{E}\big[g'(k,T,\Theta_k)\big|\Theta_{k} \big]\nonumber\\
&\le \left\|\left.\frac{\partial W}{\partial \theta}\right|_{\Theta_k}\right\|\cdot \left\|\mathbb{E}\!\left[g'(k,T,\Theta_k)\big|\Theta_{k} \right]\right\|\label{eq:inner1}\\
&\le c_4\|\Theta_k\|\cdot \epsilon LT \beta_k(T,\epsilon)(\|\Theta_k\|+1)\label{eq:term202}\\
&\le 2c_4 \epsilon LT\beta_k(T,\epsilon) (\|\Theta_k\|^2+1)\label{eq:term2}.
\end{align}
In particular, \eqref{eq:inner1} uses the Cauchy-Schwartz inequality, 
\eqref{eq:term202} calls for Proposition \ref{prop:exist}, and the last one follows from the inequality $\|\theta\|(\|\theta\|+1)\le 2(\|\theta\|^2+1)$.

As far as the last term of \eqref{eq:vdiff} is concerned, it is clear that
\begin{align}
&\quad \,\,\mathbb{E}\bigg\{\!\left[\epsilon T \wbar{f}(\Theta_k) +g'(k,T,\Theta_k)\right]^\top \!\nabla^2 W(\Theta_k') \left[\epsilon T \wbar{f}(\Theta_k) +g'(k,T,\Theta_k)\right]\!\Big|\Theta_{k} \bigg\}\nonumber\\
&\le c_4\,\mathbb{E}\!\left[ \!\left\|
\epsilon T \wbar{f}(\Theta_k) +g'(k,T,\Theta_k)\right\|^2\Big|\Theta_{k} \right]\label{eq:last1}\\
&\le 2c_4  \epsilon^2 T^2 \!\left\|\wbar{f}(\Theta_k)\right\|^2+2c_4\mathbb{E}\!\left[\big\|g'\!\left(k,T,\Theta_k\right)\big\|^2\Big|\Theta_{k}\right]
\label{eq:last2}\\
&\le 2c_4 \epsilon^2 T^2 L^2 \|\Theta_k\|^2 +2c_4\mathbb{E}\Big[\big\|g'(k,T,\Theta_k)\big\|^2\Big|\Theta_{k} \Big]\label{eq:last3}
\end{align}
where \eqref{eq:last1} leverages the upper bound on the Hessian matrix of $W(\theta)$ arising from the property  \eqref{eq:lyap3}, \eqref{eq:last2} follows from the inequality $\|a+b\|^2\le 2(\|a\|^2+\|b\|^2)$ for any real-valued vectors $a,b\in\mathbb{R}^d$, and \eqref{eq:last3} uses the Lipschitz property of function $\wbar{f}(\theta)$ that can be easily verified since $f(\theta,x)$ is Lipschitz in $\theta$. 

To further upper bound the last term of \eqref{eq:last3}, we establish the following helpful result whose proof is also postponed to Appendix \ref{pf:pgsq} for readability.

\begin{lemma}\label{le:gsqu}
	The following bound holds 
	for any fixed $\theta_k\in\mathbb{R}^d$
	\begin{align}\label{eq:gsqu}
	\mathbb{E}\!\left[
	\big\|g'(k,T,\Theta_k)\big\|^2\big| \Theta_{k}\right]
	\le \epsilon^2L^2T^2\! \left[\epsilon^2 L^2T^2 \!\left(1+\epsilon L\right)^{2T-4}+12\right]\!
	\|\Theta_k\|^2+8\epsilon^2 L^2 T^2.
	\end{align}
\end{lemma}

Coming back to inequality 
\eqref{eq:last3}, 
Lemma \ref{le:gsqu} holds true. 
Plugging \eqref{eq:gsqu} into \eqref{eq:last3},  
we establish an upper bound on the last term of \eqref{eq:vdiff} as follows
\begin{align}
&\mathbb{E}\bigg\{\!\left[\epsilon T \wbar{f}(\Theta_k) +g'(k,T,\Theta_k)\right]^\top \!\nabla^2 W(\Theta_k') \left[\epsilon T \wbar{f}(\Theta_k) +g'(k,T,\Theta_k)\right]\!\Big|\Theta_{k}\bigg\}\nonumber\\
&\le 2c_4 \epsilon^2 T^2 L^2 \left[\epsilon^2 L^2T^2 \!\left(1+\epsilon L\right)^{2T-4}+13\right] \|\Theta_k\|^2+16c_4 \epsilon^2 L^2 T^2\label{eq:lastbound}.
\end{align} 

Putting together the bounds in \eqref{eq:term1}, \eqref{eq:term2}, and \eqref{eq:lastbound}, it follows from \eqref{eq:taylor} that
\begin{align}
&\quad\,\, \mathbb{E}\!\left[W(\Theta_{k+T}(k,\Theta_k))-W(\Theta_k) \big|\Theta_{k}\right]\nonumber\\
&=\mathbb{E}\!\left[\epsilon T \!\left(\left.\frac{\partial W}{\partial \theta}\right|_{\Theta_k}\right)^\top\! \wbar{f}(\Theta_k) +\!\left(\left.\frac{\partial W}{\partial \theta}\right|_{\Theta_k}\right)^\top\! g'(k,T,\Theta_k) \Big|\Theta_{k}
\right]\nonumber\\
&\quad\, + \mathbb{E}\!\left\{\!\left[\epsilon T \wbar{f}(\Theta_k) +g'(k,T,\Theta_k)\right]^\top \!\nabla^2 W(\Theta_k') \! \left[\epsilon T \wbar{f}(\Theta_k) +g'(k,T,\Theta_k)\right]\! \Big|\Theta_{k}
\right\}\nonumber\\
&\le -\epsilon LT\!\left\{ c_3-2c_4   \beta_k(T,\epsilon)-2c_4\epsilon LT\!\left[\epsilon^2L^2T^2\!\left(1+\epsilon L\right)^{2T-4}+13 \right] \right\}\|\Theta_k\|^2\nonumber\\
&\quad \, 
+2c_4 \epsilon LT\beta_k(T,\epsilon)+16c_4 \epsilon^2 L^2 T^2\nonumber\\
&= -\epsilon LT[c_3-c_4 \rho_k(T,\epsilon)]\|\Theta_k\|^2+ c_4\epsilon LT \kappa_k(T,\epsilon)\label{eq:final}
\end{align} 
where in the last equality, we have defined for notational brevity the following two functions
\begin{align}
\rho_k(T,\epsilon)&:=2\beta_k(T,\epsilon)+2\epsilon LT\!\left[\epsilon^2L^2T^2\!\left(1+\epsilon L\right)^{2T-4}+13\right ]\label{eq:rho}\\
\kappa_k(T,\epsilon)&:=2\beta_k(T,\epsilon)+16\epsilon L T\label{eq:kappa}
\end{align}
both of which depend on parameters $T\in\mathbb{N}^+$ and $\epsilon>0$. 

In the sequel, we will show that there exist parameters $\epsilon>0$ and $T\ge 1$ such that the coefficient of \eqref{eq:final} obeys
$	
c_3-c_4 \rho_k(T,\epsilon)>0
$ for all $k\in\mathbb{N}^+$. Formally, such a result is summarized in Proposition \ref{prop:func} below, whose proof is relegated to Appendix \ref{pf:delta}.

\begin{proposition}
	\label{prop:func}
	Consider functions $\beta_k(T,\epsilon)$ and $\rho_k(T,\epsilon)$ defined in \eqref{eq:delt} and \eqref{eq:rho}, respectively. 
	Then for any $\delta>0$, there exist constants $\epsilon_{\delta}>0$ and $T_{\delta}\ge 1$, such that the following inequality holds for each $\epsilon\in (0,\,\epsilon_{\delta})$
	\begin{equation}
	\label{eq:rhobound}
	\sigma(T_\delta,k)< \rho_k(T_{\delta},\epsilon)<\rho_0(T_{\delta},\epsilon)<  \rho_0(T_{\delta},\epsilon_{\delta})\le \delta,\quad \forall k\ge 1. 
	\end{equation}
\end{proposition}

As such, by taking any $\delta<{c_3}/{c_4}$, feasible parameter values $T^\ast$ and $\epsilon_c$ can be obtained according to \eqref{eq:tdelta} and \eqref{eq:epsilonc}, respectively. Now by choosing 
\begin{align}
\label{eq:tstar}
T^\ast &= T_\delta\\
\epsilon_{c}&=\epsilon_\delta\label{eq:epsc}
\end{align}
it follows
that
\begin{equation}\label{eq:c3prime}
c_3':=LT^\ast \left[c_3-c_4\rho_0(T^\ast,\epsilon_{\delta})\right]=LT^\ast \left(c_3-c_4\delta\right)>0.
\end{equation}  


It follows from \eqref{eq:final} that
\begin{align}
\mathbb{E}\!\left[W(\Theta_{k+T}(k,\Theta_k))-W(\Theta_k) \big|\Theta_{k}
\right]&\le -c_3' \epsilon  \|\Theta_k\|^2+c_4\epsilon LT^\ast \kappa_k(T^\ast,\epsilon)
\nonumber\\
&=-c_3' \epsilon  \|\Theta_k\|^2+
c_4'\epsilon^2 + c_5' \sigma(T^\ast;k) \epsilon\label{eq:mid1}
\end{align}
where we have defined constants $c_4':=c_4LT^\ast\big[2L(1+\epsilon_{\delta} L)^{T^\ast-2}+16LT^\ast \big]$, and $c_5':=2c_4LT^\ast $.

Finally, recalling \eqref{eq:vdiff}, we deduce that
\begin{equation}
\mathbb{E}\!\left[W'(k+1,\Theta_k+\epsilon f(\Theta_k,X_k))-	W'(k,\Theta_k) \big|\Theta_{k}
\right]\le  -c_3' \epsilon \|\Theta_k\|^2+c_4'\epsilon^2 + c_5' \sigma(T^\ast;k) \epsilon
\end{equation}
concluding the proof of \eqref{eq:exis2}.	

\vspace{.5cm}
Now, we turn to show the first inequality. 
It is evident from the properties of $W(\Theta_k)$ in Assumption \ref{as:lyap} that
\begin{align}\label{eq:vprime}
W'(k,\Theta_k)=\sum_{j=k}^{k+T-1} W(\Theta_j(k,\Theta_k))
&\ge W(\Theta_k(k,\Theta_k))\nonumber\\
&\ge c_1\|\Theta_k(k,\Theta_k)\|^2\nonumber\\
&=c_1\|\Theta_k\|^2
\end{align}
where the second inequality follows from \eqref{eq:lyap1}, and the last equality from the fact that $\Theta_k(k,\Theta_k)=\Theta_k$. Therefore, by taking $c_1'=c_1$, we have shown that
the first part of inequality \eqref{eq:exis1} holds true.  For the second part, it follows that
\begin{align}
\|\Theta_{j+1}\|=\|\Theta_j+\epsilon f(\Theta_j,X_j)\|\le (1+\epsilon L)\|\Theta_j\|+ \epsilon L,\quad \forall j\ge k
\end{align}
yielding by means of telescoping series
\begin{align}
\|\Theta_j(k,\Theta_k)\|&\le (1+\epsilon L)^{j-k}\|\Theta_k\|+\sum_{j=1}^{j-k}(1+\epsilon L)^{j-1}\epsilon L\nonumber\\
&\le (1+\epsilon L)^{j-k}\|\Theta_k\|+(1+\epsilon L)^{j-k}-1,\quad \forall j\ge k.\nonumber
\end{align}
Using further the inequality $(a+b)^2\le 2(a^2+b^2)$, we deduce that
\begin{equation}
\label{eq:thetasquare}
\|\Theta_j(k,\Theta_k)\|^2\le 2(1+\epsilon L)^{2(j-k)}\|\Theta_k\|^2+2\!\left[\!\left(1+\epsilon L\right)^{j-k}-1 \right]^2.
\end{equation}

Taking advantage of the properties of $W(\Theta_k)$ in Assumption \ref{as:lyap} and \eqref{eq:thetasquare}, it follows that
\begin{align}
W'(k,\Theta_k)&=\sum_{j=k}^{k+T-1} W(\Theta_j(k,\Theta_k))\nonumber\\
&\le\sum_{j=k}^{k+T-1} c_2 \|\Theta_j(k,\Theta_k) \|^2\nonumber\\
&\le 2c_2   \!\sum_{j=k}^{k+T-1}(1+\epsilon L)^{2(j-k)}\|\Theta_k\|^2+2c_2\sum_{j=k}^{k+T-1}\!\left[\!\left(1+\epsilon L\right)^{j-k}-1 \right]^2 \label{eq:squarethetafinal}.
\end{align}

Let us now examine the two coefficients of \eqref{eq:squarethetafinal} more carefully. Note that
\begin{align}
\sum_{j=k}^{k+T-1}(1+\epsilon L)^{2(j-k)}&=\frac{(1+\epsilon L)^{2T}-1}{(1+\epsilon L)^2-1}=T\, \frac{2+(2T-1)(1+\epsilon'L)^{2T-2}\epsilon L}{2+\epsilon L}
\label{eq:c2prime11}\\
\sum_{j=k}^{k+T-1}\!\left[(1+\epsilon L)^{j-k}-1 \right]^2
&= \sum_{j=k+1}^{k+T-1}\left[(j-k) \epsilon L \left(1+\frac{1}{2}(j-k-1)\!\left(1+\epsilon_{j-k}' L\right)^{j-k-2}\epsilon L
\right)
\right]^2\label{eq:c2prime12}
\\
&= (\epsilon L)^2 \sum_{j=1}^{T-1} j^2 \left[1+\frac{1}{2} (j-1)\!\left(1+\epsilon_j' L\right)^{j-2}
\right]^2
\label{eq:c2primep13}
\end{align}
where both \eqref{eq:c2prime11} and \eqref{eq:c2prime12} follow from the mean-value theorem $(1+\epsilon L)^{j-k}=1 +(j-k) \epsilon L +\frac{1}{2} (j-k-1)(1+\epsilon_{j-k} 'L)^{j-k-2}(\epsilon L)^2 $ for any $j-k\ge 1$ and some constants $\epsilon_j' \in [0,\epsilon]$. 

According to Proposition \ref{prop:func}, or more specifically, the inequalities \eqref{eq:tstar} and \eqref{eq:epsc}, we see that
$\epsilon_j'\le \epsilon\le \epsilon_{\delta}$ for all $1\le j\le T-1$. 

On the other hand, it is easy to check that both terms [\eqref{eq:c2prime11} and \eqref{eq:c2primep13}] are monotonically increasing functions of $\epsilon>0$. 
Therefore, if we define constants 
\begin{align}
c_2'&:=2c_2  T^\ast\, \frac{2+(2T^\ast -1)(1+\epsilon_{\delta}L)^{2T^\ast-2}\epsilon_{\delta} L}{2+\epsilon_{\delta} L} 
\label{eq:c2prime}\\
c_2''&:=2c_2\sum_{j=1}^{T^\ast-1} j^2 \left[1+\frac{1}{2} (j-1)\!\left(1+\epsilon_{\delta} L\right)^{j-2}
\right]^2
\label{eq:c2primep}
\end{align}
which are independent of $\epsilon$, 
then we draw from \eqref{eq:squarethetafinal}, \eqref{eq:c2prime11}, and \eqref{eq:c2primep13} that
\begin{equation}
W'(k,\Theta_k)\le c_2' \|\Theta_k\|^2+ c_2'' (\epsilon L)^2 \label{eq:secondpart}.
\end{equation}
concluding the proof of the second part of \eqref{eq:exis1}.


\section{Proof of Lemma \ref{le:recursive}}

\label{pf:recursive}
Taking expectation of both sides of \eqref{eq:exis1} conditioned on $\Theta_k$ gives rise to
\begin{equation}
\mathbb{E}\big[W'(k,\Theta_k)|	\Theta_k
\big]
\le c_2' \|\Theta_k\|^2+c_2''(\epsilon L)^2.\label{eq:relax0}
\end{equation}
On the other hand, it is evident from \eqref{eq:exis2} that
\begin{align}
&\quad\; \mathbb{E}\big[W'(k+1,\Theta_{k+1})| 
\Theta_k\big]\nonumber\\
& \le \mathbb{E}\big[W'(k,\Theta_k)| 
\Theta_k\big] -c_3' \epsilon \|\Theta_k\|^2+c_4'\epsilon^2 + c_5' \sigma(T^\ast;k) \epsilon
 \nonumber\\
&= \mathbb{E}\big[W'(k,\Theta_k)| 	\Theta_k\big] -\frac{c_3'\epsilon}{c_2'}\!  \left[c_2'
\|\Theta_k\|^2+c_2'' (\epsilon L)^2 \right]+\frac{c_3'}{c_2'}c_2''\epsilon (\epsilon L)^2+c_4'\epsilon^2 + c_5' \sigma(T^\ast;k) \epsilon\nonumber\\
&\le  \mathbb{E}\big[W'(k,\Theta_k)| 	\Theta_k\big] -\frac{c_3'\epsilon}{c_2'}   \mathbb{E}\big[W'(k,\Theta_k)| 	\Theta_k\big]+\frac{c_3'}{c_2'}c_2''\epsilon_{\delta} (\epsilon L)^2 +c_4'\epsilon^2 + c_5' \sigma(T^\ast;k) \epsilon \label{eq:relax1}\\
&=\left( 1-\frac{c_3'\epsilon}{c_2'}\right)\mathbb{E}\big[W'(k,\Theta_k)| 	\Theta_k\big] +c_4''\epsilon^2 +c_5' \sigma(T^\ast;k) \epsilon
\label{eq:condlast}
\end{align}
where, in order to obtain \eqref{eq:relax1}, we have employed the inequality in \eqref{eq:relax0},
 and used the fact that $\epsilon<\epsilon_{\delta}$ to derive \eqref{eq:condlast}; and the last equality follows from $c_4'':=c_4'+{c_3'c_2''\epsilon_{\delta} L^2}/{c_2'}$.

Finally, taking expectation of both sides of \eqref{eq:condlast} with respect to $\Theta_k$, concludes the proof.


\section{Proof of Theorem \ref{th:finite}}
\label{pf:finite}

Let us start with a basic Lemma, whose proof is elementary and is hence omitted here.

\vspace{.18cm}
\begin{lemma}
	Consider the recursion $z_{t+1}=az_t+b$, where $a\ne 1$ and $b$ are given constants. Then the following holds for all $t\ge t_0\ge 0$
	\begin{equation}\label{eq:rec}
	z_{t}=a^{t-t_0} z_{t_0}+\frac{b(a^{t-t_0}-1)}{a-1}.
	\end{equation}
\end{lemma}

Proof of Theorem \ref{th:finite} is established in two phases depending on the $k$ values. Specifically, let us define $k_{\epsilon}:=\min\{k\in\mathbb{N}^+|\sigma(T^\ast;k)\le \epsilon  \}$; then the first phase is from $k=0$ to $k_\epsilon$, while the second phase consists of all $k>k_\epsilon$.

\vspace{.5cm}
\noindent\emph{Phase I ($k\le k_\epsilon$)}. We have from \ref{prop:func} that $\sigma(T^\ast;k)\le \delta$ for all $0\le k(\le k_\epsilon)$. Then, fixing $t_0=0$, and substituting $a:=1- {c_3'\epsilon}/{c_2'}>0$ and $b:=c_4''\epsilon^2 +c_5' \delta \epsilon $ in \eqref{eq:rec}, the recursion $\{\mathbb{E}[W'(k,\Theta_k)]\}$ in \eqref{eq:vdrift} can be recursively expressed as follows
\begin{align}
\mathbb{E}\big[W'(k,\Theta_{k})
\big]&\le \left(1- \frac{c_3'\epsilon}{c_2'} \right) \mathbb{E}\big[W'(k-1,\Theta_{k-1})
\big]+c_4''\epsilon^2 +c_5' \delta \epsilon\nonumber\\
&\le \left(1- \frac{c_3'\epsilon}{c_2'} \right)^{\!k}\! \mathbb{E}\big[W'(0,\Theta_0) 
\big]
+\bigg[1-\left(1-\frac{c_3'\epsilon}{c_2'}
\right)^{\! k}
\bigg]\frac{ c_2'}{c_3'}\big(c_4''\epsilon+c_5'\delta\big)\label{eq:mid3}\\
&\le \left(1- \frac{c_3'\epsilon}{c_2'} \right)^{\!k}\! \mathbb{E}\big[W'(0,\Theta_0) 
\big]
+\frac{ c_2'}{c_3'}\big(c_4''\epsilon+c_5'\delta\big) \nonumber\\
&\le \left(1- \frac{c_3'\epsilon}{c_2'} \right)^{\!k}\! \mathbb{E}\big[W'(0,\Theta_0) 
\big]
+\frac{ c_2'}{c_3'}\big(c_4''+c_5'\big)\delta\label{eq:mid33}\\
&\le c_2' \left(1- \frac{c_3'\epsilon}{c_2'} \right)^{\!k} \|\Theta_0\|^2+c_2''L^2\epsilon^2+
c_6\delta
\label{eq:wpright}
\end{align}
where the last inequality follows from $\epsilon\le \delta$ and the fact [cf. \eqref{eq:exis1}] that 
\begin{equation}
\mathbb{E}[W'(0,\Theta_0) 
]\le c_2'\mathbb{E}[\|\Theta_0\|^2]+c_2''\epsilon^2 L^2\le c_2'\|\Theta_0\|^2+c_2''\epsilon^2 L^2\label{eq:zero}
\end{equation}
where the initial guess $\Theta_0\in\mathbb{R}^d$ is assumed given for simplicity; and $c_6:=c_2(c_4''+c_5')/c_3'$.


On the other hand, using \eqref{eq:exis1}, 
the term $\mathbb{E}\big[W'(k,\Theta_{k})
\big]$ can be lowered bounded as follows
\begin{equation}
\mathbb{E}\big[W'(k,\Theta_{k})
\big]\ge c_1'\|\Theta_k\|^2\label{eq:llower}
\end{equation}
which, combined with \eqref{eq:wpright}, yields the finite-time error bound for iterations $k\le k_\epsilon$
\begin{equation}
\mathbb{E}[\|\Theta_k\|^2]\le \frac{c_2'}{c_1'} \left(1- \frac{c_3'\epsilon}{c_2'} \right)^{\!k} \|\Theta_0\|^2+\frac{c_2'' L^2}{c_1'}\epsilon^2
 +\frac{c_6}{c_1'}\delta.
\end{equation}

\vspace{.5cm}
\noindent\emph{Phase II ($k>k_\epsilon$)}. 
Using now the fact that $\sigma(T^\ast;k)\le \epsilon$ due to the definition of $k_\epsilon$, the recursion $\{\mathbb{E}[W'(k,\Theta_k)]\}$ for all $k>k_\epsilon$ becomes
\begin{align}
		\mathbb{E}\big[W'(k+1,\Theta_{k+1})
\big]&\le \left(1- \frac{c_3'\epsilon}{c_2'} \right) \mathbb{E}\big[W'(k,\Theta_k)
\big]+ c_4''\epsilon^2 +c_5' \sigma(T^\ast;k) \epsilon\\
&\le \left(1- \frac{c_3'\epsilon}{c_2'} \right) \mathbb{E}\big[W'(k,\Theta_k)
\big]+ (c_4''+c_5')\epsilon^2 
\label{eq:rewrite}.
\end{align}

Letting $t_0=k_\epsilon$, and replacing $a$ and $b$ in \eqref{eq:rec} by constants $(1- {c_3'\epsilon}/{c_2'})$ and $(c_4''+c_5')\epsilon^2 $ accordingly, we arrive at
\begin{align}
\mathbb{E}\big[W'(k,\Theta_{k})
\big]
&\le \left(1- \frac{c_3'\epsilon}{c_2'} \right)^{\!k-k_\epsilon}\! \mathbb{E}\big[W'(k_\epsilon,\Theta_{k_\epsilon}) 
\big]
+\bigg[1-\left(1-\frac{c_3'\epsilon}{c_2'}
\right)^{\! k-k_\epsilon}
\bigg]\frac{ c_2'}{c_3'}\big(c_4''+c_5'\big)\epsilon\nonumber\\
&\le \left(1- \frac{c_3'\epsilon}{c_2'} \right)^{\!k-k_\epsilon}\!\left[ \left(1- \frac{c_3'\epsilon}{c_2'} \right)^{\!k_\epsilon}\! \mathbb{E}\big[W'(0,\Theta_0) 
\big]
+\frac{ c_2'}{c_3'}\big(c_4''\epsilon+c_5'\delta\big)\right]
+\frac{ c_2'(c_4''+c_5')}{c_3'}\epsilon \nonumber\\
&\le \left(1- \frac{c_3'\epsilon}{c_2'} \right)^{\!k}\mathbb{E}\big[W'(0,\Theta_0) 
\big]+ \left(1- \frac{c_3'\epsilon}{c_2'} \right)^{\!k-k_\epsilon}\frac{ c_2'\big(c_4''+c_5'\big)}{c_3'}\delta
+\frac{ c_2'(c_4''+c_5')}{c_3'}\epsilon
\label{eq:mid4}\\
&\le c_2'\!\left(1- \frac{c_3'\epsilon}{c_2'} \right)^{\!k}\|\Theta_0\|^2
+c_2''\epsilon^2L^2+
 \!\left(1- \frac{c_3'\epsilon}{c_2'} \right)^{\!k-k_\epsilon}c_6
 \delta
+c_6
\epsilon\label{eq:finn}
\end{align}
where we have used the following bound at $k=k_\epsilon$ from Phase I in \eqref{eq:mid3} along with \eqref{eq:zero}
\begin{equation}
\mathbb{E}[W'(k_{\epsilon},\Theta_{k_\epsilon}) ]\le \left(1- \frac{c_3'\epsilon}{c_2'} \right)^{\!k_\epsilon}\! \mathbb{E}\big[W'(0,\Theta_0) 
\big]
+\bigg[1-\left(1-\frac{c_3'\epsilon}{c_2'}
\right)^{\! k_\epsilon}
\bigg]\frac{ c_2'}{c_3'}\big(c_4''\epsilon+c_5'\delta\big).
\end{equation}

Plugging \eqref{eq:llower} into \eqref{eq:finn}, yields the finite-time error bound for $k\ge k_\epsilon$
\begin{align}
\label{eq:finalphase2}
\mathbb{E}\big[\|\Theta_k\|^2\big]\le \frac{c_2'}{c_1'}\!\left(1- \frac{c_3'\epsilon}{c_2'} \right)^{\!k}\!\|\Theta_0\|^2+\frac{c_2'' L^2}{c_1'}\epsilon^2+ \!\left(1- \frac{c_3'\epsilon}{c_2'} \right)^{\!k-k_\epsilon}\!\!\frac{ c_6}{c_1'}\delta
+\frac{ c_6}{c_1'}\epsilon
\end{align}
which converges to a small (size-$\epsilon$) neighborhood of the optimal solution $\Theta^\ast=0$ at a linear rate. 

Combining the results in the two phases, we deduce the following error bound that holds at any $k\in\mathbb{N}^+$
\begin{align}
\mathbb{E}\big[\|\Theta_k\|^2\big]\le \frac{c_2'}{c_1'}\!\left(1- \frac{c_3'\epsilon}{c_2'} \right)^{\!k}\!\|\Theta_0\|^2+\frac{c_2'' L^2}{c_1'}\epsilon^2+ 
\!\left(1- \frac{c_3'\epsilon}{c_2'} \right)^{\max\{k-k_\epsilon,0\}}\!\!\frac{ c_6}{c_1'}\delta
+\frac{ c_6}{c_1'}\epsilon
\end{align}
concluding the proof of Theorem \ref{th:finite}.

\section{Proof of Lemma \ref{le:gbound}}
\label{pf:le1}
\vspace{.18cm}

When $T=1$ and for any $\Theta_k\in\mathbb{R}^d$, one can easily check that 
\begin{equation*}
g(k,1,\Theta_k)=\Theta_{k+1}-\Theta_k-\epsilon f(\Theta_k,X_{k})=0
\end{equation*}
implying $G_1:=\|g(k,1,\Theta_k)\|=0$. 
To proceed, let us start by introducing the function $$h(k,T,\Theta_k):=\sum_{j=k}^{k+T-1} f(\Theta_k,X_{j})$$
which can be bounded as follows
\begin{align}
\big\| h(k,T,\Theta_k)\big\|
=\!\left\| \sum_{j=k}^{k+T-1}  \! f(\Theta_k,X_j)\right\|
&\le \sum_{j=k}^{k+T-1}
\big \| f(\Theta_k,X_j)\big\|\nonumber\\
&\le L  \sum_{j=k}^{k+T-1}(\|\Theta_k\|+1)\nonumber\\
&=TL (\|\Theta_k\|+1)\label{eq:hnorm}
\end{align}
where the second inequality follows from \eqref{eq:fbound} in Assumption \ref{as:fxy}.

It is evident that
\begin{align}
g(k,T+1,\Theta_k)&=\Theta_{k+T+1}-\Theta_k-\epsilon \sum_{j=k}^{k+T} f(\Theta_k,X_{j} )\nonumber\\
&=\Theta_{k+T}+\epsilon f(\Theta_{k+T},X_{k+T})-\Theta_k- \epsilon \bigg[f(\Theta_k,X_{k+T_0} )+\sum_{j=k}^{k+T-1} f(\Theta_k,X_{j} )\bigg] \nonumber\\
&=g(k,T,\Theta_k)+\epsilon\big[f(\Theta_{k+T},X_{k+T})-f(\Theta_k,X_{k+T} )\big].
\end{align}
By means of triangle inequality, it follows that
\begin{align}
G_{T+1}=\|g(k,T+1,\Theta_k)\|
&\le \big\|g(k,T,\Theta_k)\big\|+\epsilon\big \| f(\Theta_{k+T},X_{k+T})-f(\Theta_k,X_{k+T} )\big\|\nonumber\\
&\le G_T+\epsilon L\big \|\Theta_{k+T}-\Theta_k\big\|\label{eq:gt13}\\
&\le G_T+\epsilon L\left [\epsilon\big\|h(k,T,\Theta_k)\big\|+\big\|g(k,T,\Theta_k)\big\| \right]\label{eq:gt14}\\
&\le (1+\epsilon L) G_T+\epsilon^2 L^2T(\|\Theta_k\|+1)\label{eq:gt15}\\
&\le 
\epsilon^2L^2(\|\Theta_k\|+1)\sum_{k=0}^{T}(1+\epsilon L)^{T-k} k
\label{eq:gt}
\end{align}
where the inequality \eqref{eq:gt13} follows from the Lipschitz continuity of $f(\theta,x)$ in $\theta$, \eqref{eq:gt14} from the fact that  $\Theta_{k+T}=\Theta_k+\epsilon h(k,T,\Theta_k)+g(k,T,\Theta_k)$, \eqref{eq:gt15} from \eqref{eq:hnorm} as well as the definition $G_{T}:=\|g(k,T,\Theta_k)\|$,
and the last inequality is obtained by telescoping series and uses $G_1=0$.

\begin{lemma}
	\label{le:prod}
	Given any positive constant $d\ne 1$, the following holds for all $T\ge 1$
	\begin{equation}
	S_{T+1}=\sum_{k=0}^{T}kd^k =\frac{d(1-d^T)}{(1-d)^2}-\frac{Td^{T+1}}{1-d}.\label{eq:prod}
	\end{equation}
\end{lemma}

Taking $d=(1+\epsilon L)^{-1}$ in \eqref{eq:prod}, then \eqref{eq:gt} can be simplified as follows
\begin{align}
G_{T}&\le \epsilon^2L^2(1+\epsilon L)^{T-1}(\|\Theta_k\|+1)\sum_{k=0}^{T-1}(1+\epsilon L)^{-k} k\nonumber\\
&=\left[ (1+\epsilon L)^{T}-\epsilon LT-1\right](\|\Theta_k\|+1).\label{eq:gt0}
\end{align} 

To further simplify this bound, the Taylor expansion along with the mean-value theorem confirms that the following holds for some $\epsilon'\in (0,1)$
\begin{equation}\label{eq:expa}
(1+\epsilon L)^{T}=1+\epsilon L T+\frac{1}{2} T(T-1) (1+\epsilon' L)^{T-2} {(\epsilon L)}^2,\quad \forall T\ge 1
\end{equation}
or equivalently,
\begin{align}
(1+\epsilon L)^{T}-1-\epsilon L T&=\frac{1}{2} T(T-1) (1+\epsilon' L)^{T-2} {(\epsilon L)}^2\label{eq:second0}\\
&\le\epsilon^2L^2 T^2(1+\epsilon L)^{T-2}.\label{eq:second}
\end{align}

\section{Proof of Lemma \ref{le:gsqu}}
\label{pf:pgsq}

Recalling that $g'(k,T,\Theta_k)=g(k,T,\Theta_k)+\epsilon\sum_{j=k}^{k+T-1}[ f(\Theta_k,X_{j})-\wbar{f}(\Theta_k) ]$, we have
\begin{align}
\big\|g'(k,T,\Theta_k)\big\|^2&=
\bigg \| g(k,T,\Theta_k)+\epsilon\sum_{j=k}^{k+T-1}\!\left( f(\Theta_k,X_{j})-\wbar{f}(\Theta_k) \right)\bigg\|^2\nonumber\\
&\le 2\,
\big\|g(k,T,\Theta_k)\big\|^2 + 2\epsilon^2T^2 \, \bigg\|\frac{1}{T}\! \sum_{j=k}^{k+T-1} f(\Theta_k,X_{j})-\wbar{f}(\Theta_k)\bigg\|^2\label{eq:gsqu2}\\
&\le 4 \left[ (1+\epsilon L)^{T}-\epsilon LT-1\right]^2   (\|\Theta_{k}\|^2+1)\nonumber\\
&\quad  + 4\epsilon^2T^2\,
\bigg\|\frac{1}{T}\! \sum_{j=k}^{k+T-1}  f(\Theta_k,X_j)\bigg\|^2+ 4\epsilon^2T^2\, \big\| \wbar{f}(\Theta_k) \big\|^2  
\label{eq:gsqu3}
\end{align}
where we have used the property $\|a+b\|^2\le 2(\|a\|^2+\|b\|^2)$ for any real-valued vectors $a,b$ in deriving \eqref{eq:gsqu2} and \eqref{eq:gsqu3}, as well as Proposition \ref{prop:exist}.

Squaring both sides of \eqref{eq:second0} yields
\begin{equation}
\left[(1+\epsilon L)^T-1-\epsilon TL\right]^2=\frac{1}{4} T^2(T-1)^2 (\epsilon L)^4 (1+\epsilon' L)^{2T-4}\le \frac{1}{4} \epsilon^4 L^4T^4 (1+\epsilon L)^{2T-4}.
\end{equation}
Thus, the first term of \eqref{eq:gsqu3} can be upper bounded by 
\begin{equation}
4 \left[ (1+\epsilon L)^{T}-\epsilon LT-1\right]^2  (\|\Theta_k\|^2+1)\le \epsilon^4 L^4T^4 (1+\epsilon L)^{2T-4} \, (\|\Theta_k\|^2 +1).\label{eq:term11}
\end{equation}

Regarding the second term of \eqref{eq:gsqu3}, we have that
\begin{align}
\bigg\|\frac{1}{T}\! \sum_{j=k}^{k+T-1}  f(\Theta_k,X_j)\bigg\|^2
&\le \frac{1}{T} \sum_{j=k}^{k+T-1} \big\| f(\Theta_k,X_j)\big\|^2
\label{eq:term22}\\
&\le \frac{1}{T}\! \sum_{j=k}^{k+T-1}  \!L^2(\|\Theta_k\|+1)^2
\label{eq:term23}\\
&\le 2L^2\|\Theta_k\|^2 + 2L^2\label{eq:term24}
\end{align}
where \eqref{eq:term22} and \eqref{eq:term24} follow from the inequality $\|\sum_{i=1}^n z_i\|^2\le n\sum_{i=1}^n \|z_i\|^2$ for all real-valued vectors $\{ z_i\}_{i=1}^n$, and \eqref{eq:term23} from our working assumption on function $f(\theta,x)$.

Withe regards to the last term of \eqref{eq:gsqu3}, it follows directly from the Lipschitz property of the average operator $\wbar{f}(\theta)$ that
\begin{equation}
\left\| \wbar{f}(\Theta_k) \right\|^2
\le L^2 \|\Theta_k\|^2.
\label{eq:term31}
\end{equation}

Substituting the bounds in \eqref{eq:term11}, \eqref{eq:term24}, and \eqref{eq:term31}
into \eqref{eq:gsqu3}, we arrive at
\begin{align}
\left\|g'(k,T,\Theta_k)\right\|^2&\le \epsilon^2L^2T^2\!\left[\epsilon^2 L^2T^2 \!\left(1+\epsilon L\right)^{2T-4}+12\right]\! \|\Theta_k\|^2+8\epsilon^2 L^2 T^2
\end{align}
concluding the proof.

\section{Proof of Proposition \ref{prop:func}}
\vspace{.18cm}
\label{pf:delta}
We prove this claim by construction. By definition, it follows that for all $k\in\mathbb{N}^+$
\begin{equation}
\rho_k(T,\epsilon)\le \rho_0(T,\epsilon)=2 \epsilon LT\!\left[\left (1+\epsilon L\right)^{T-2}+13
\right]+2(\epsilon L T)^3(1+\epsilon L)^{2T-4}+2\sigma(T;0).
\end{equation} 
Under the assumption that $\lim_{T\to+\infty}\sigma(T;0)=0$, the function value $\sigma(T;0)\ge 0$ can be made arbitrarily small by taking a sufficiently large integer $T\in\mathbb{N}^+$ in constructing the function $W'(k,\Theta_k)$. Without loss of generality, let us work with $T$ such that
\begin{equation}\label{eq:tdelta}
T_{\delta}:=\min \left\{T\in \mathbb{N}^+\left|\sigma(T;0)\le \frac{\delta}{4} \right. \right \}.
\end{equation}

It is clear that $T_{\delta}\ge 1$. Define function 
\begin{equation}
\label{eq:firstwo}
\nu(\epsilon):= \epsilon LT_{\delta}\!\left[ \left(1+\epsilon L\right)^{T_{\delta}-2}+13
\right]+\!\left(\epsilon L T_{\delta}\right)^3(1+\epsilon L)^{2T_{\delta}-4}
\end{equation}
which can be easily shown to be a monotonically decreasing function of $\epsilon>0$, and which attains its minimum $\nu=0$ at $\epsilon=0$. 
Let $\epsilon_{\delta}$ be the unique solution to the equation 
\begin{equation}
\label{eq:epsilonc}
\nu(\epsilon)=\frac{\delta}{4},\quad \epsilon>0.
\end{equation}
As a result, for all $\epsilon\in (0,\epsilon_{\delta}]$, it holds that
\begin{equation}\label{eq:nueps}
\nu(\epsilon)\le \frac{\delta}{4}.
\end{equation}

Combining \eqref{eq:tdelta} and \eqref{eq:nueps} concludes the proof of Proposition \ref{prop:func}.

%
%


\end{document}